\documentclass{article}

\PassOptionsToPackage{numbers,sort&compress}{natbib}
\usepackage[main, final]{neurips_2026}

\usepackage[utf8]{inputenc} 
\usepackage[T1]{fontenc}    
\usepackage{hyperref}       
\usepackage{url}            
\usepackage{booktabs}       
\usepackage{amsfonts}       
\usepackage{nicefrac}       
\usepackage{microtype}      
\usepackage[dvipsnames]{xcolor}         

\usepackage{amsmath}
\usepackage{amssymb}

\usepackage{wrapfig}
\usepackage{pifont}
\newcommand{\xmark}{\ding{55}}
\newcommand{\cmark}{\ding{51}}
\usepackage{graphicx}
\usepackage{multirow}
\usepackage{bm}
\usepackage{adjustbox}

\usepackage{chngcntr}
\usepackage{marvosym}

\usepackage{subcaption}

\usepackage{algorithm}
\usepackage[noEnd=true,indLines=true,rightComments=false,italicComments=false,
commentColor=teal,beginLComment=//~,endLComment=]{algpseudocodex}

\tikzset{
    phaseduring/.style={draw=none, fill=ForestGreen!15},
    phaseafter/.style={draw=none, fill=Orange!18},
    algpxIndentLine/.style={draw=black, thin}
}

\DeclareRobustCommand{\correspondingmark}{%
  \ensuremath{\text{\Letter}}%
}

\makeatletter
\let\original@fnsymbol\@fnsymbol
\renewcommand{\@fnsymbol}[1]{%
  \ifnum#1=1\relax
    \correspondingmark
  \else
    \original@fnsymbol{#1}%
  \fi
}
\makeatother

\usepackage{makecell}

\title{Modulating Merging Strengths via Joint Loss Estimation for LoRA-based Continual Learning}

\author{%
\begin{tabular}{c}
\textbf{Kun Gu}\textsuperscript{1}
\quad
\textbf{De Cheng}\textsuperscript{1,}\thanks{\normalfont Corresponding author}
\quad
\textbf{Zhipeng Xu}\textsuperscript{1}
\quad
\textbf{Lingfeng He}\textsuperscript{1}
\quad
\textbf{Di Xu}\textsuperscript{2}
\quad
\textbf{Nannan Wang}\textsuperscript{1}
\\[4pt]
{\normalfont
\textsuperscript{1}Xidian University
\qquad
\textsuperscript{2}Huawei Technologies Co., Ltd.}
\\[2pt]
\footnotesize
\texttt{\{kung@stu., dcheng@, xu\_zhipeng@stu., lfhe@stu., nnwang@\}xidian.edu.cn}
\\
\footnotesize
\texttt{xudi21@huawei.com}
\end{tabular}%
}

\begin{document}

\maketitle


\begin{abstract}
Continual learning (CL) aims to adapt models to new tasks while preserving previously learned knowledge.
Recently, Low-Rank Adaptation (LoRA), a representative Parameter-Efficient Fine-Tuning (PEFT) method, has gained increasing attention in CL due to its efficiency and scalability.
Several LoRA-based merging methods maintain a single inference model by scaling the newly learned LoRA update with one coefficient and merging it into the previous weights after each task.
However, they usually control the whole update with a shared scalar factor, although different update elements may affect previous and current tasks in different ways.
To address this issue, we propose \underline{M}erging Strength \underline{M}odulation via Joint \underline{L}oss \underline{E}stimation ($\text{M}^2$LE).
It consists of Hessian Information-guided Adaptive Merging (HIM) and Task Simulation Perturbation for Merging (TSP).
Instead of searching for a global merge coefficient, HIM directly solves for the target merged update under a joint objective over previous and current tasks.
Solving this objective gives an analytic solution that naturally takes the form of element-wise modulation over the current task LoRA update.
To improve the stability of the learned update under subsequent merging, we introduce TSP which simulates different tasks interference.
Experiments on multiple benchmarks show that $\text{M}^2$LE consistently outperforms existing PEFT-based continual learning methods. Code is available at \href{https://github.com/bbdeeer/M2LE_NeurIPS2026}{\textcolor{blue}{https://github.com/bbdeeer/M2LE\_NeurIPS2026}}.
\end{abstract}
\section{Introduction}

Continual learning (CL) aims to enable models to learn continuously from a stream of tasks without retraining from scratch or accessing data from previous tasks~\cite{ewc,cl1}, which is essential for evolving open-world scenarios.
A central challenge in CL is catastrophic forgetting: models must preserve previously learned capabilities during adapting to new tasks~\cite{cl3,li2026few}.
In this work, we consider the challenging rehearsal-free Continual Learning setting, where the model must continually learn new tasks without storing past data and predict test samples over all seen tasks without task identities.

Recently, with the advancement of Pre-Trained Models (PTMs)~\cite{ptm}, Parameter-Efficient Fine-Tuning (PEFT) methods have emerged as a promising solution for CL~\cite{l2p,codaprompt,ssiat}. By updating a small number of paramters, PEFT-based methods enable efficient adaptation to new tasks while preserving general pre-trained knowledge, making it particularly suitable for rehearsal-free CL.

In the regime of PEFT-based CL, Low-Rank Adaptation (LoRA)~\cite{lora} methods have gained increasing attention due to their efficiency and scalability.
To maintain a single inference model, a common strategy is to merge the current task LoRA update into the accumulated model after each task~\cite{pm}, rather than storing a separate LoRA module per task.
This controlled merging reduces interference with prior knowledge while enabling adaptation to the current task.
Early studies mainly adopt EMA-style interpolation~\cite{ema} or weight averaging~\cite{acmap}, 
while recent approaches estimate the interpolation coefficient from joint objectives~\cite{pm} or Bayesian formulations~\cite{became}.
Despite these different designs, most existing methods still use a shared scalar factor to control the whole LoRA update.
This naturally raises a question: \textbf{\emph{should all LoRA parameters share the same merge strength, or should different positions in the LoRA update be treated differently during merging?}}
To answer this question, we conduct a preliminary analysis in Fig.~\ref{fig:pre_analysis}, 
where the LoRA update is split into two groups according to different normalized magnitudes~\cite{rapf}.
We then assign an independent merge coefficient to each group.
The results show that this simple grouped merge already outperforms the best shared factor merge, suggesting that different parts of the LoRA update may prefer different merge strengths.

Building on this observation, 
we propose a \underline{M}erging Strength \underline{M}odulation via Joint \underline{L}oss \underline{E}stimation ($\text{M}^2$LE) framework for LoRA-based Continual Learning, consists of Hessian Information-guided Adaptive Merging (HIM) and Task Simulation Perturbation for Merging (TSP).
Instead of searching for an interpolation coefficient, HIM directly formulates the optimization of the target merged update $\Delta \mathbf{W}_t^*$ under a joint objective that accounts for losses on all seen tasks. 
Through a local quadratic approximation to the task-wise losses, we obtain an analytic solution, which naturally takes the form of an element-wise modulation of the current task LoRA update. 
Specifically, each element is adaptively scaled by the associated Hessian information, allowing the merging process to account for the varying sensitivity of individual parameter elements across tasks, thereby achieving a more fine-grained balance between old and new task performance.
However, once the current task update is merged into the inference model, it becomes part of the fixed parameters and will be affected by subsequent task updates.
If the merged update lies in a sharp local region, later merging steps can easily increase the loss on old tasks, resulting in forgetting.
Therefore, besides determining how to merge the update, it is also important to shape the update itself before merging.
To this end, we introduce TSP, which simulates cross-task interference during current task training.
TSP induces a local parameter change with one subset of the current mini-batch and evaluates the resulting loss variation on another subset, thereby encouraging the learned update to remain stable during subsequent merging.

Our main contributions can be summarized as follows:
\begin{itemize}
\item 
We propose Hessian Information-guided Adaptive Merging (HIM), a parameter-adaptive merging algorithm derived from an approximated joint loss objective, enabling more fine-grained stability-plasticity balancing beyond a shared scalar merging coefficient.

\item 
We propose Task Simulation Perturbation (TSP), a training strategy that simulates future cross-task interference through local parameter perturbations induced from label-disjoint subset, thereby encouraging flat solutions and improving LoRA merging stability.
\item 
Extensive experiments demonstrate the effectiveness of our proposed method, achieving  superior performances over state-of-the-art PEFT-based CL methods on multiple benchmarks.

\end{itemize}

\section{Related Work}



\textbf{PEFT-based and LoRA-based continual learning.} PEFT-based CL~\cite{SIKD} typically keeps a pre-trained model frozen and introduces lightweight learnable modules for continual learning.
L2P~\cite{l2p} retrieves instance specific prompts from a prompt pool, and SECA~\cite{seca} uses textual semantic priors to support knowledge transfer and prototype refinement in CLIP-based continual learning.
EWC~\cite{ewc} preserves previous knowledge by penalizing changes to important parameters, and EKPC~\cite{ekpc} mitigates forgetting through elastic knowledge preservation and compensation.
Among PEFT-based method~\cite{DPR,PAPT,PADG,RD-MLDG,PAPT++,chenginterference, wang2026stpr,DCD}, LoRA has become a representative choice for continual learning due to its parameter efficiency and compatibility with frozen pre-trained backbones~\cite{lora}. 
InfLoRA~\cite{inflora} constrains new LoRA updates to interference-free subspaces through orthogonality, PLAN~\cite{plan} proactively allocates low-rank subspaces and optimizes them against worst-case perturbations, 
GainLoRA~\cite{gainlora} expands a new LoRA for each task and introduces gating modules to adaptively weight the old and new branches, 
and MACIL~\cite{macil} mitigates forgetting through semantic drift calibration. LoDA~\cite{loda2026} continually decomposes LoRA updates into general and isolated subspaces for better stability-plasticity trade-off.



\textbf{Model Merging.} Model merging aims to integrate weights trained on different tasks into a single network~\cite{merge1}. 
RobustMerge~\cite{robustmerge} improves parameter-efficient merging by enhancing the robustness of low-rank update directions, while OPCM~\cite{opcm} enables sequential merging through orthogonal projection and adaptive scaling. 
This idea has also been used in continual learning, where the current model or update is merged into the inference model after each task~\cite{rapf}.
ACMap~\cite{acmap} averages the adapters learned from different tasks into a shared adapter, 
BECAME~\cite{became} derives an adaptive coefficient from a Bayesian formulation, and LoRA-P\&M~\cite{pm} combines the previous and current models using a shared merging factor computed in closed form.
However, these methods mainly rely on a shared factor to interpolate  updates, while our method derives the merged update from the loss changes of old and new tasks under a joint objective.
\section{Preliminary Analysis}
\label{pre_analysis}


\textbf{Research Question.}
\emph{Should LoRA parameters at different positions be treated differently during merging, rather than being controlled by a shared factor?}

\textbf{Experiment Setting.}
As a case study, we consider the 10 task setting on ImageNet-R and focus on the first merge step. Specifically, we take the model after learning task 0 and the update learned on task 1 for analysis. 
Let $\mathbf{W}_{0}$ denote the inference parameter after learning task 0, and let $\Delta \mathbf{W}_1$ denote the LoRA update learned on task 1.
To this end, we build a simple merging setup. Specifically, we compute the normalized magnitude map of $\Delta \mathbf{W}_{1}$ as $\mathbf{R}=|\Delta \mathbf{W}_{1}| / \max_{i,j} |\Delta \mathbf{W}_{1}^{(i,j)}|$~\cite{rapf}, and divide the update into two groups: elements with normalized values below 0.5, and elements with normalized values at least 0.5. 
We then assign two independent coefficients, $\alpha_1$ and $\alpha_2$, to the two groups, and define the merged model as: $\mathbf{W}(\alpha_1,\alpha_2)=\mathbf{W}_{0}+\alpha_1(\mathbf{M}_1\odot\Delta \mathbf{W}_{1})+\alpha_2(\mathbf{M}_2\odot\Delta \mathbf{W}_{1})$, where $\mathbf{M}_1$ and $\mathbf{M}_2$ denote the corresponding masks, and $\odot$ denotes element-wise multiplication. 
By varying $\alpha_1$ and $\alpha_2$, we evaluate the joint loss on task 0 and task 1.

\begin{wrapfigure}{r}{0.4\textwidth}
    \centering
    \vspace{-6.0mm}
    \includegraphics[width=\linewidth]{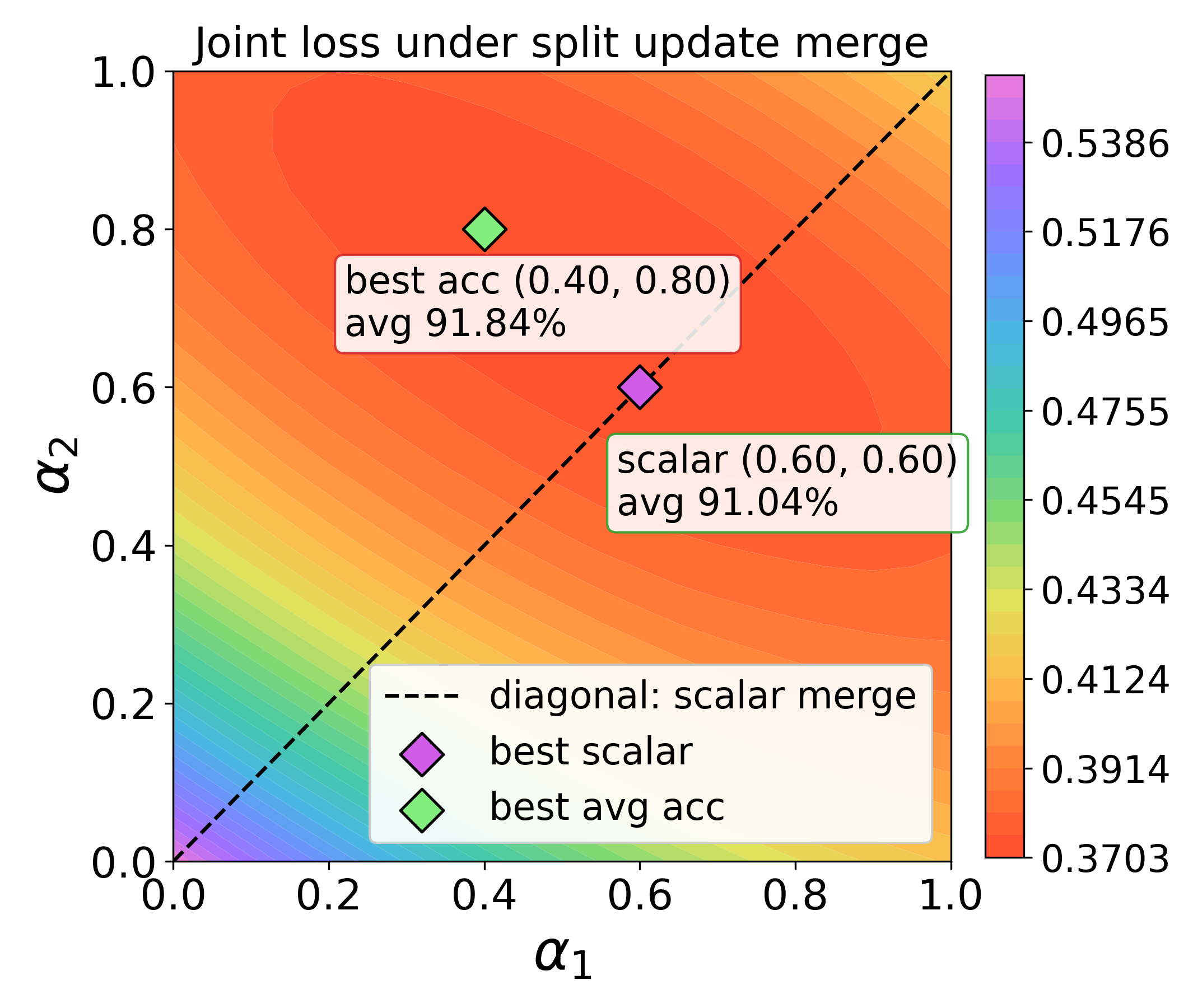}
    \vspace{-6.0mm}
    \caption{Joint loss landscape of the two group merge on 10S-ImageNet-R.}
    \label{fig:pre_analysis}
    \vspace{-6.0mm}
\end{wrapfigure}

\textbf{Observation.}
Fig.~\ref{fig:pre_analysis} shows that the low loss region is not centered on the diagonal $\alpha_1=\alpha_2$, but is shifted toward an off-diagonal region. 
Consistently, the \textcolor{ForestGreen}{green diamond} marks the point with the best average accuracy on task0 and task1, which is located at $(0.40, 0.80)$, while the \textcolor{purple}{purple diamond} marks the best scalar merge on the diagonal, located at $(0.60, 0.60)$.
Compared with the scalar merge, the green-diamond point improves the average accuracy from $91.04\%$ to $91.84\%$.

\textbf{Insight.}
These results suggest that a shared scalar coefficient is overly restrictive: different parameter groups do not, in general, share the same preferred merge strength, and should therefore be treated differently during merging.

\section{Methodology}

\subsection{Preliminaries}
In CL, a model learns from a sequence of tasks
$\{\mathcal{D}_t\}_{t=1}^{T}$.
The dataset of task $t$ is denoted by
$\mathcal{D}_t=\{(x_t^i,y_t^i)\}_{i=1}^{n_t}$,
where $x_t^i\in\mathbb{R}^{d}$ is the input and
$y_t^i\in\mathcal{C}_t$ is the corresponding label.
Each task introduces a set of new classes, and the label spaces of different
tasks are mutually disjoint, i.e.,
$\mathcal{C}_i\cap\mathcal{C}_j=\varnothing$ for any $i\neq j$.
Accordingly, the cumulative label space observed up to task $t$ is
$\mathcal{C}_{1:t}=\bigcup_{i=1}^{t}\mathcal{C}_i$. 
Under the standard rehearsal-free CL setting, only the current task data
$\mathcal{D}_t$ is accessible when learning task $t$.
The goal is to learn tasks sequentially while maintaining strong performance
over all tasks observed so far.

\subsection{Overall framework}
Let $\mathbf{W}_{t-1}\in \mathbb{R}^{d_{o}\times d_{i}}$ denote the inference parameter after learning task $t-1$, where $d_{i}$ and $d_{o}$ are the input and output dimensions.
At $t$-th task, for each ViT layers, we freeze the backbone weights $\mathbf{W}_{t-1}$ and attach a LoRA module.
The LoRA update is parameterized by two low-rank matrices:
\begin{equation}
\Delta \mathbf{W}_t = \mathbf{B}_t\mathbf{A}_t,
\quad
\mathbf{A}_t\in\mathbb{R}^{r\times d_i},\quad
\mathbf{B}_t\in\mathbb{R}^{d_o\times r},
\label{eq:lora_update}
\end{equation}
where $r$ is the LoRA rank. 
A common LoRA merging strategy directly updates the inference model by scaling the current task update with a shared coefficient $\mathbf{W}_t = \mathbf{W}_{t-1} + \alpha_t \Delta \mathbf{W}_t$,
where the scalar $\alpha_t$ controls the merge strength of the whole update.
However, this scalar merge applies the same merge strength to all elements in $\Delta \mathbf{W}_t$, although different elements may have different effects on previous and current tasks.

To address this, $\text{M}^2$LE consists of two components: Hessian Information-guided Adaptive Merging (HIM) and Task Simulation Perturbation for Merging (TSP).
During training, TSP regularizes the LoRA update by simulating cross task perturbations, encouraging the learned update $\Delta \mathbf{W}_t$ to improve the stability during subsequent merging.
After training, HIM derives a target merged update ${\Delta \mathbf{W}_t^{*}}$ under a joint loss impact objective over previous and current tasks, instead of searching for a global merge coefficient.
The inference parameter is then updated by $\mathbf{W}_t = \mathbf{W}_{t-1} + {\Delta \mathbf{W}_t^{*}}$.
An overview of the framework and the complete training procedure
are provided in Appendix~\ref{training}
(Figure~\ref{fig:framework} and Algorithm~\ref{alg:liam}).



\subsection{Hessian Information-guided Adaptive Merging}
Existing LoRA-based merging methods for CL always merge the current task LoRA update into the inference model using a shared scalar factor.
This forces all parameter elements to be merged with the same strength, although they may affect previous and current tasks differently.
This motivates us to account for the different preferred merge strengths of update parameter during merging.

Specifically, we directly solve for the target merged update $\Delta {\mathbf{W}_t^*}\in \mathbb{R}^{d_{o}\times d_{i}}$ under a joint objective over previous and current tasks.
Intuitively, we want this update to incorporate new knowledge as effectively as possible while reducing catastrophic forgetting on previous tasks.
To this end, we first define $\mathcal{J}_{\mathrm{old}}$ and $\mathcal{J}_{\mathrm{new}}$ to measure how an ideal merged update $\hat{{\Delta \mathbf{W}}_t}\in \mathbb{R}^{d_{o}\times d_{i}}$ affects the losses on previous and current tasks:
\begin{equation}
\mathcal{J}_{\mathrm{old}}(\hat{{\Delta \mathbf{W}}_t})
=
\sum_{i=1}^{t-1}
\left(
\mathcal{L}_{i}(\mathbf{W}_{t-1}+\hat{{\Delta \mathbf{W}}_t})
-
\mathcal{L}_{i}(\mathbf{W}_{t-1})
\right),
\label{eq:old_task_term}
\end{equation}
\begin{equation}
\mathcal{J}_{\mathrm{new}}(\hat{{\Delta \mathbf{W}}_t})
=
\mathcal{L}_{t}(\mathbf{W}_{t-1}+\hat{{\Delta \mathbf{W}}_t})
-
\mathcal{L}_{t}(\mathbf{W}_{t-1}+\Delta \mathbf{W}_t).
\label{eq:new_term}
\end{equation}

Here, $\Delta \mathbf{W}_t\in \mathbb{R}^{d_{o}\times d_{i}}$ denotes the update learned for task $t$, $\mathcal{L}_i$ denotes the loss of task $i$.
Therefore, we seek a merged update that jointly minimizes $\mathcal{J}_{\mathrm{old}}$ on previous tasks and $\mathcal{J}_{\mathrm{new}}$ on the current task, with a hyper-parameter $\lambda$ controlling the strength of the current task term.
The target merged update ${\Delta \mathbf{W}_t^{*}}$ is obtained by solving:
\begin{equation}
{\Delta \mathbf{W}_t^*}
=
\arg\min_{\hat{{\Delta \mathbf{W}}_t}}
\;
\mathcal{J}_{\mathrm{old}}(\hat{{\Delta \mathbf{W}}_t})
+
\lambda \mathcal{J}_{\mathrm{new}}(\hat{{\Delta \mathbf{W}}_t}).
\label{eq:ideal_obj}
\end{equation}

In the rehearsal-free setting, Eq.~\eqref{eq:ideal_obj} cannot be optimized directly, since the old-task term $\mathcal{J}_{\mathrm{old}}$ depends on previous task data that are no longer accessible when learning task $t$.
To obtain a tractable approximation, we approximate Eq.~\eqref{eq:old_task_term} and Eq.~\eqref{eq:new_term} with local second-order Taylor expansions.
For notational simplicity, 
we vectorize 
$\Delta \hat{ \mathbf{W}}_t$
into $\Delta {\hat{ \mathbf{w}}_t}=\mathrm{vec}(\Delta\hat{ \mathbf{W}}_t)\in\mathbb{R}^{d_od_i \times 1}$.
For the old task term $\mathcal{J}_{\mathrm{old}}(\hat{{\Delta \mathbf{w}}_t})$, expanding around $
\mathbf{w}_{t-1}$ gives:
\begin{equation}
\mathcal{J}_{\mathrm{old}}(\hat{{\Delta \mathbf{w}}_t})
\approx
\sum_{i=1}^{t-1}
\left(
\mathbf{g}_i^\top \hat{{\Delta \mathbf{w}}_t}
+
\frac{1}{2}\hat{{\Delta \mathbf{w}}_t}^\top \mathbf{H}_i \hat{{\Delta \mathbf{w}}_t}
\right),
\label{eq:taylor_old}
\end{equation}
where $\mathbf{g}_i \in\mathbb{R}^{d_od_i\times 1}$ and $\mathbf{H}_i \in\mathbb{R}^{d_od_i \times d_od_i}$ denote the first and second terms of the $i$-th old task loss around $\mathbf{w}_{t-1}$, respectively.
Similarly, for the current task, expanding around $\mathbf{w}_{t-1}+\Delta \mathbf{w}_t$ yields:

\begin{equation}
\mathcal{J}_{\mathrm{new}}(\hat{{\Delta \mathbf{w}}_t})
\approx
\mathbf{g}_t^\top (\hat{{\Delta \mathbf{w}}_t}-\Delta \mathbf{w}_t)
+
\frac{1}{2}(\hat{{\Delta \mathbf{w}}_t}-\Delta \mathbf{w}_t)^\top \mathbf{H}_t(\hat{{\Delta \mathbf{w}}_t}-\Delta \mathbf{w}_t),
\label{eq:taylor_new}
\end{equation}
where $\mathbf{g}_t$ and $\mathbf{H}_t$ denote the first- and second-order terms of the current task loss around $\mathbf{w}_{t-1}+\Delta \mathbf{w}_t$, respectively.
Since the previous model $\mathbf{w}_{t-1}$ is obtained by recursively
merging past updates and $\mathbf{w}_{t-1}+\Delta\mathbf{w}_t$ is the learned
current-task solution.
Under a local approximation, we treat the first-order terms in
Eq.~\eqref{eq:taylor_old} and \eqref{eq:taylor_new} as approximately zero.
Substituting the remaining quadratic terms into
Eq.~\eqref{eq:ideal_obj}, we obtain:

\begin{equation}
\begin{aligned}
\Delta \mathbf{W}_t^*
= \arg\min_{\hat{{\Delta \mathbf{w}}_t}} \quad
& \sum_{i=1}^{t-1} \frac{1}{2}\hat{{\Delta \mathbf{w}}_t}^{\top} \mathbf{H}_i \hat{{\Delta \mathbf{w}}_t} \\
& + \frac{\lambda}{2}
(\hat{{\Delta \mathbf{w}}_t}-\Delta \mathbf{w}_t)^{\top}
\mathbf{H}_t
(\hat{{\Delta \mathbf{w}}_t}-\Delta \mathbf{w}_t).
\end{aligned}
\label{eq:quad_obj_main}
\end{equation}

Taking the derivative with respect to ${\Delta \hat{\mathbf{w}}}_t$ and setting it to zero gives:
\begin{equation}
\left(\sum_{i=1}^{t-1} \mathbf{H}_i+\lambda \mathbf{H}_t \right) \Delta \mathbf{w}_t^*
=
\lambda \mathbf{H}_t \Delta \mathbf{w}_t.
\label{eq:closed_form_main}
\end{equation}
A detailed derivation is provided in Appendix~\ref{merge}, and Appendix~\ref{fisher} further analyzes the effectiveness
of accumulated historical Fisher statistics.
Following EWC~\cite{ewc}, we approximate each Hessian in Eq.~\eqref{eq:closed_form_main} with the diagonal empirical Fisher, reusing their historical running sum for the old task term.
For task $t$, the $i$-th diagonal element of $\mathbf{F}_t$
at the learned parameters $\mathbf{W}_{t-1}+\Delta\mathbf{W}_t$
is defined as $
\mathbf{F}_t^{i,i}
=
\mathbb{E}_{(x,y)\sim\mathcal{D}_t}
\left[
\left(
\left.
\frac{\partial \mathcal{L}(f_{\mathbf{W}}(x),y)}
{\partial w_i}
\right|_{\mathbf{W}=\mathbf{W}_{t-1}+ \Delta \mathbf{W}_t}
\right)^2
\right]$.
Here, $\mathbf{F}_t\in\mathbb{R}^{d_o\times d_i}$ denotes the diagonal Fisher of the trainable update, reshaped to match $\Delta\mathbf{W}_t$.
Under this diagonal approximation, Eq.~\eqref{eq:closed_form_main} gives the following element-wise merged update:
\begin{equation}
\Delta \mathbf{W}_t^*
=
\frac{\lambda \mathbf{F}_t}
{
\mathbf{F}_{1:t-1}
+\lambda\mathbf{F}_t+\varepsilon}
\odot
\Delta \mathbf{W}_t,
\label{eq:element_wise_solution}
\end{equation}
where $\varepsilon$ is a small damping constant for numerical stability and after merging task $t$, we update the accumulated Fisher by
$\mathbf{F}_{1:t}=\mathbf{F}_{1:t-1}+\mathbf{F}_t$, which is used in the next merging step.

\subsection{Task Simulation Perturbation for Merging}

According to Eq.~\eqref{eq:element_wise_solution}, HIM obtains $\Delta \mathbf{W}_t^*$ by element-wise reweighting $\Delta \mathbf{W}_t$ and merge it into the inference model, where it becomes part of the fixed parameters for later tasks.
If the merged update lies in a sharp region, subsequent new task merging can easily increase its loss and lead to forgetting.
This naturally connects to the flat-minimum principle, making SAM~\cite{sam} a straightforward way to improve the local stability of the current task update.
However, SAM computes the perturbation from the overall current task loss and minimizes the perturbed loss on the same task, so its flatness constraint remains limited to the current task.
Such a design makes it difficult to approximate the case in CL where the update from one task affects another task.
To approximating cross-task interference when old task data are unavailable. 
We randomly split each mini-batch into two label-disjoint subsets $(\mathcal{S}_1,\mathcal{S}_2)$, using one subset to induce a local parameter change and the other to evaluate the resulting loss change.

For the $\mathcal{S}_1 \rightarrow \mathcal{S}_2$ direction, we first compute the gradient on the
source subset $\mathcal{S}_1$:
\begin{equation}
\mathbf{g}_1 = \nabla \mathcal{L}(\mathcal{S}_1).
\label{eq:meta_source_grad}
\end{equation}
We then construct a probing perturbation along the induced descent direction:
\begin{equation}
\mathbf{\epsilon}_1 = -\rho \frac{\mathbf{g}_1}{\|\mathbf{g}_1\|_2^2+\varepsilon},
\label{eq:meta_probe}
\end{equation}
where $\rho$ controls the perturbation scale and $\varepsilon$ is a small constant for numerical stability. 
The target subset $\mathcal{S}_2$ is evaluated under both the perturbed and original parameters:
\begin{equation}
\mathcal{L}_{\mathrm{pert}}^{1\rightarrow 2}
=
\mathcal{L}(\mathcal{S}_2; \mathbf{B}_t\mathbf{A}_t+\mathbf{\epsilon}_1),\qquad
\mathcal{L}_{\mathrm{clean}}^{1\rightarrow 2}
=
\mathcal{L}(\mathcal{S}_2; \mathbf{B}_t\mathbf{A}_t).
\label{eq:meta_losses}
\end{equation}
The perturbed loss evaluates whether the direction induced by $\mathcal{S}_1$ remains effective on the disjoint subset $\mathcal{S}_2$ under a local displacement. 
To prevent the perturbation from pulling training too far away from the normal update and degrading the clean loss on $\mathcal{S}_2$, we retain the clean loss as an anchor and define the TSP objective as:
\begin{equation}
\mathcal{L}_{\mathrm{TSP}}^{1\rightarrow 2}
=
\beta \mathcal{L}_{\mathrm{pert}}^{1\rightarrow 2}
+
(1-\beta)\mathcal{L}_{\mathrm{clean}}^{1\rightarrow 2},
\label{eq:meta_mix}
\end{equation}
where $\beta$ balances the perturbed and clean terms.
We first update the parameters for the $\mathcal{S}_1 \rightarrow \mathcal{S}_2$ step by:
\begin{equation}
\mathbf{B}'_t\mathbf{A}'_t
=
\mathbf{B}_t\mathbf{A}_t
-
\eta \nabla \mathcal{L}_{\mathrm{TSP}}^{1\rightarrow 2}(\mathbf{B}_t\mathbf{A}_t),
\label{eq:meta_update_ab}
\end{equation}
where $\eta$ is the learning rate.
We then exchange the roles of the two subsets and perform the reverse update:
\begin{equation}
\mathbf{B}''_t\mathbf{A}''_t
=
\mathbf{B}'_t\mathbf{A}'_t
-
\eta \nabla \mathcal{L}_{\mathrm{TSP}}^{2\rightarrow 1}(\mathbf{B}'_t\mathbf{A}'_t),
\label{eq:meta_update_ba}
\end{equation}
where $\mathcal{L}_{\mathrm{TSP}}^{2\rightarrow 1}$ is defined in the same way, with the roles of $\mathcal{S}_1$ and $\mathcal{S}_2$ exchanged.
Thus, within each mini-batch, the optimization proceeds sequentially as $\mathbf{B}_t\mathbf{A}_t
\;\xrightarrow{\;\mathcal{S}_1\rightarrow \mathcal{S}_2\;}\;
\mathbf{B}'_t\mathbf{A}'_t
\;\xrightarrow{\;\mathcal{S}_2\rightarrow \mathcal{S}_1\;}\;
\mathbf{B}''_t\mathbf{A}''_t$.
This dual training framework learns the current task update under task simulating perturbations, so that the subsequent merge better preserves previously learned knowledge.

\section{Experiment}
\label{experment}

\begin{table*}[t]
\centering
\caption{
Experimental results for 10 incremental tasks on ImageNet-R, ImageNet-A, CIFAR100, and CUB. We report the averaged results over 3 random seeds. Within each group, the highest results are in \textbf{bold}, and the second highest results are \underline{underlined}. "FR" denotes the use of feature replay.}
\label{tab:main_10s}
\resizebox{\textwidth}{!}{
\begin{tabular}{l c c cc cc cc cc}
\toprule
\multirow{2}{*}{Method} & \multirow{2}{*}{Venue} & \multirow{2}{*}{FR} 
& \multicolumn{2}{c}{10S-ImageNet-R} 
& \multicolumn{2}{c}{10S-ImageNet-A}
& \multicolumn{2}{c}{10S-CIFAR100}
& \multicolumn{2}{c}{10S-CUB} \\

& & & $\mathcal{A}_{{Last}}\uparrow$ & $\mathcal{A}_{{Avg}}\uparrow$
& $\mathcal{A}_{{Last}}\uparrow$ & $\mathcal{A}_{{Avg}}\uparrow$
& $\mathcal{A}_{{Last}}\uparrow$ & $\mathcal{A}_{{Avg}}\uparrow$
& $\mathcal{A}_{{Last}}\uparrow$ & $\mathcal{A}_{{Avg}}\uparrow$ \\
\midrule

L2P~\cite{l2p} & CVPR'22 & \xmark
& ${72.34}_{\pm 0.17}$ & ${77.36}_{\pm 0.64}$
& ${44.04}_{\pm 0.93}$ & ${51.24}_{\pm 2.26}$
& ${84.06}_{\pm 0.88}$ & ${88.26}_{\pm 1.34}$
& ${67.02}_{\pm 1.90}$ & ${70.62}_{\pm 1.60}$ \\

CODAPrompt~\cite{codaprompt} & CVPR'23 & \xmark
& ${73.31}_{\pm 0.50}$ & ${78.47}_{\pm 0.53}$
& ${52.08}_{\pm 0.12}$ & ${63.92}_{\pm 0.12}$
& ${83.21}_{\pm 3.39}$ & ${87.71}_{\pm 3.17}$
& ${77.23}_{\pm 1.12}$ & ${81.90}_{\pm 0.85}$ \\

InfLoRA~\cite{inflora} & CVPR'24 & \xmark
& ${74.75}_{\pm 0.64}$ & ${80.67}_{\pm 0.55}$
& ${49.20}_{\pm 1.12}$ & ${60.92}_{\pm 0.61}$
& ${86.75}_{\pm 0.35}$ & ${91.72}_{\pm 0.15}$
& ${70.82}_{\pm 0.23}$ & ${81.39}_{\pm 0.14}$ \\

PLAN~\cite{plan} & ICCV'25 & \xmark
& ${75.25}_{\pm 0.42}$ & ${80.41}_{\pm 0.56}$
& -- & --
& ${87.54}_{\pm 0.31}$ & ${92.21}_{\pm 0.35}$
& -- & -- \\

SEMA~\cite{sema} & CVPR'25 & \xmark
& ${78.00}\phantom{_{\pm 0.00}}$ & ${83.56}\phantom{_{\pm 0.00}}$
& ${53.32}\phantom{_{\pm 0.00}}$ & ${64.53}\phantom{_{\pm 0.00}}$
& ${86.98}\phantom{_{\pm 0.00}}$ & ${91.37}\phantom{_{\pm 0.00}}$
& -- & -- \\

SD-LoRA~\cite{sdlora} & ICLR'25 & \xmark
& ${77.34}_{\pm 0.35}$ & ${82.04}_{\pm 0.24}$
& ${55.96}_{\pm 0.73}$ & ${64.95}_{\pm 1.63}$
& ${88.01}_{\pm 0.31}$ & ${92.54}_{\pm 0.18}$
& ${77.48}_{\pm 0.20}$ & $\underline{85.59}_{\pm 0.44}$ \\

BiLoRA~\cite{bilora} & CVPR'25 & \xmark
& ${77.95}_{\pm 0.14}$ & ${81.52}_{\pm 0.26}$
& -- & --
& ${87.46}_{\pm 0.76}$ & ${92.50}_{\pm 0.62}$
& -- & -- \\

LoRA-P\&M~\cite{pm} & NIPS'25 & \xmark 
& $\underline{79.95}_{\pm 0.18}$ & $\underline{85.29}_{\pm 0.93}$ 
& ${56.57}_{\pm 0.78}$ & ${65.35}_{\pm 1.81}$ 
& $\underline{88.45}_{\pm 0.35}$ & $\underline{92.89}_{\pm 1.13}$
& $\underline{78.29}_{\pm 0.50}$ & ${83.39}_{\pm 0.61}$ \\

EWC-LoRA~\cite{ewc-lora} & ICLR'26 & \xmark 
& ${72.86}_{\pm 0.79}$ & ${78.95}_{\pm 0.86}$ 
& $\underline{59.89}_{\pm 0.26}$ & $\underline{68.33}_{\pm 0.67}$ 
& ${87.91}_{\pm 0.57}$ & ${92.27}_{\pm 0.39}$
& --  & -- \\

\midrule
$\textbf{$\textbf{M}^2$LE (Ours)}$ & -- & \xmark 
& $\mathbf{81.34}_{\bm{\pm}\mathbf{0.16}}$ & $\mathbf{86.53}_{\bm{\pm}\mathbf{0.51}}$ 
& $\mathbf{62.45}_{\bm{\pm}\mathbf{1.34}}$ & $\mathbf{71.02}_{\bm{\pm}\mathbf{1.82}}$ 
& $\mathbf{89.97}_{\bm{\pm}\mathbf{0.04}}$ & $\mathbf{93.31}_{\bm{\pm}\mathbf{0.95}}$
& $\mathbf{81.40}_{\bm{\pm}\mathbf{0.21}}$ & $\mathbf{89.23}_{\bm{\pm}\mathbf{0.48}}$ \\
\midrule

SLCA~\cite{slca} & ICCV'23 & \cmark
& ${79.35}_{\pm 0.28}$ & ${83.29}_{\pm 0.46}$
& ${61.05}_{\pm 0.63}$ & ${68.88}_{\pm 2.31}$
& ${91.26}_{\pm 0.37}$ & ${94.29}_{\pm 0.92}$
& ${84.68}_{\pm 0.09}$ & ${90.77}_{\pm 0.79}$ \\

SSIAT~\cite{ssiat} & CVPR'24 & \cmark 
& ${79.38}_{\pm 0.59}$ & ${83.63}_{\pm 0.43}$ 
& ${62.43}_{\pm 1.63}$ & ${70.83}_{\pm 1.63}$ 
& ${91.35}_{\pm 0.26}$ & ${94.35}_{\pm 0.60}$
& ${88.75}_{\pm 0.38}$ & ${93.00}_{\pm 0.90}$ \\

MOS~\cite{mos} & AAAI'25 & \cmark
& ${77.68}_{\pm 0.41}$ & ${82.06}_{\pm 0.53}$
& ${54.75}_{\pm 1.09}$ & ${63.32}_{\pm 2.38}$
& ${91.53}_{\pm 0.35}$ & ${94.21}_{\pm 0.91}$
& ${89.97}_{\pm 0.32}$ & ${93.43}_{\pm 0.60}$ \\

MACIL~\cite{macil} & ICML'25 & \cmark 
& $\underline{81.88}_{\pm 0.07}$ & $\underline{85.95}_{\pm 0.27}$ 
& ${64.14}_{\pm 0.58}$ & ${71.45}_{\pm 1.35}$ 
& $\mathbf{91.94}_{\bm{\pm}\mathbf{0.17}}$ & ${94.43}_{\pm 0.79}$
& $\mathbf{90.52}_{\bm{\pm}\mathbf{0.13}}$ & $\mathbf{93.93}_{\bm{\pm}\mathbf{0.47}}$ \\

SAEF~\cite{saef} & CVPR'26 & \cmark 
& ${80.02}\phantom{_{\pm 0.00}}$ & ${84.54}\phantom{_{\pm 0.00}}$ 
& $\underline{64.91}\phantom{_{\pm 0.00}}$ & $\underline{73.39}\phantom{_{\pm 0.00}}$ 
& ${90.60}\phantom{_{\pm 0.00}}$ & $\underline{94.53}\phantom{_{\pm 0.00}}$
& -- & -- \\

\midrule
$\textbf{$\textbf{M}^2$LE (Ours) + CA}$ & -- & \cmark
& $\mathbf{82.68}_{\bm{\pm}\mathbf{0.30}}$ & $\mathbf{87.05}_{\bm{\pm}\mathbf{0.51}}$ 
& $\mathbf{65.96}_{\bm{\pm}\mathbf{0.90}}$ & $\mathbf{74.02}_{\bm{\pm}\mathbf{0.85}}$ 
& $\underline{91.86}_{\pm 0.21}$ & $\mathbf{94.62}_{\bm{\pm}\mathbf{0.90}}$
& $\underline{90.35}_{\pm 0.16}$ & $\underline{93.92}_{\pm 0.38}$ \\
\bottomrule
\end{tabular}
}
\end{table*}

\begin{table*}[t]
\centering
\caption{Additional results under different task settings, including 5 and 20 tasks on ImageNet-R, 20 tasks on ImageNet-A, and 10 tasks on DomainNet. Within each group, the highest results are in \textbf{bold}, and the second highest results are \underline{underlined}. ``FR'' denotes the use of feature replay.}
\label{tab:main_table_2}
\resizebox{\textwidth}{!}{
\begin{tabular}{l c c cc cc cc cc}
\toprule
\multirow{2}{*}{Method} & \multirow{2}{*}{Venue} & \multirow{2}{*}{FR} 
& \multicolumn{2}{c}{5S-ImageNet-R} 
& \multicolumn{2}{c}{20S-ImageNet-R}
& \multicolumn{2}{c}{20S-ImageNet-A}
& \multicolumn{2}{c}{10S-DomainNet} \\

& & & $\mathcal{A}_{{Last}}\uparrow$ & $\mathcal{A}_{{Avg}}\uparrow$
& $\mathcal{A}_{{Last}}\uparrow$ & $\mathcal{A}_{{Avg}}\uparrow$
& $\mathcal{A}_{{Last}}\uparrow$ & $\mathcal{A}_{{Avg}}\uparrow$
& $\mathcal{A}_{{Last}}\uparrow$ & $\mathcal{A}_{{Avg}}\uparrow$ \\
\midrule

L2P~\cite{l2p} & CVPR'22 & \xmark
& ${70.83}_{\pm 0.58}$ & ${78.34}_{\pm 0.47}$
& ${69.64}_{\pm 0.42}$ & ${75.28}_{\pm 0.57}$
& ${40.48}_{\pm 1.78}$ & ${49.62}_{\pm 1.46}$
& ${81.17}_{\pm 0.83}$ & ${87.43}_{\pm 0.95}$ \\

CODAPrompt~\cite{codaprompt} & CVPR'23 & \xmark
& ${74.91}_{\pm 0.33}$ & ${79.25}_{\pm 0.53}$
& ${69.96}_{\pm 0.50}$ & ${75.34}_{\pm 0.85}$
& ${44.62}_{\pm 1.92}$ & ${54.86}_{\pm 0.50}$
& ${80.04}_{\pm 0.79}$ & ${86.27}_{\pm 0.82}$ \\

InfLoRA~\cite{inflora} & CVPR'24 & \xmark
& ${77.52}_{\pm 0.37}$ & ${82.01}_{\pm 0.12}$
& ${71.01}_{\pm 0.45}$ & ${77.28}_{\pm 0.45}$
& ${46.81}_{\pm 2.30}$ & ${57.13}_{\pm 0.95}$
& ${81.45}_{\pm 0.68}$ & ${87.55}_{\pm 0.57}$ \\

SEMA~\cite{sema} & CVPR'25 & \xmark
& ${79.78}\phantom{_{\pm 0.00}}$ & ${84.75}\phantom{_{\pm 0.00}}$
& ${74.53}\phantom{_{\pm 0.00}}$ & ${81.75}\phantom{_{\pm 0.00}}$
& -- & --
& -- & -- \\

SD-LoRA~\cite{sdlora} & ICLR'25 & \xmark
& ${79.15}_{\pm 0.20}$ & ${83.01}_{\pm 0.42}$
& ${75.26}_{\pm 0.37}$ & ${80.22}_{\pm 0.72}$
& ${49.12}_{\pm 1.78}$ & ${59.63}_{\pm 1.57}$
& ${81.01}_{\pm 0.42}$ & ${86.85}_{\pm 0.40}$ \\

BiLoRA~\cite{bilora} & CVPR'25 & \xmark
& ${78.32}_{\pm 0.37}$ & ${83.31}_{\pm 0.52}$
& ${72.41}_{\pm 0.65}$ & ${79.28}_{\pm 0.76}$
& -- & --
& ${76.56}_{\pm 0.41}$ & ${83.20}_{\pm 0.35}$ \\

LoRA-P\&M~\cite{pm} & NIPS'25 & \xmark  
& $\underline{81.47}_{\pm 0.56}$ & $\underline{85.96}_{\pm 0.52}$ 
& $\underline{76.37}_{\pm 0.09}$ & $\underline{82.77}_{\pm 0.71}$ 
& $\underline{52.27}_{\pm 1.84}$ & $\underline{62.15}_{\pm 1.33}$
& $\underline{84.35}_{\pm 0.48}$ & $\underline{89.43}_{\pm 0.27}$ \\

EWC-LoRA~\cite{ewc-lora} & ICLR'26 & \xmark 
& ${76.36}_{\pm 0.21}$ & ${81.43}_{\pm 0.13}$ 
& ${70.18}_{\pm 1.06}$ & ${77.06}_{\pm 0.54}$ 
& -- & --
& --  & -- \\

\midrule
$\textbf{$\textbf{M}^2$LE (Ours)}$ & -- & \xmark 
& $\mathbf{82.67}_{\bm{\pm}\mathbf{0.18}}$ & $\mathbf{87.31}_{\bm{\pm}\mathbf{0.35}}$
& $\mathbf{78.62}_{\bm{\pm}\mathbf{0.43}}$ & $\mathbf{84.62}_{\bm{\pm}\mathbf{0.26}}$
& $\mathbf{53.87}_{\bm{\pm}\mathbf{1.45}}$ & $\mathbf{63.69}_{\bm{\pm}\mathbf{1.87}}$
& $\mathbf{84.78}_{\bm{\pm}\mathbf{0.13}}$ & $\mathbf{89.74}_{\bm{\pm}\mathbf{0.44}}$ \\
\midrule

SLCA~\cite{slca} & ICCV'23 & \cmark
& ${81.01}_{\pm 0.11}$ & ${84.18}_{\pm 0.28}$
& ${74.63}_{\pm 1.55}$ & ${79.92}_{\pm 1.29}$
& ${36.69}_{\pm 21.31}$ & ${56.35}_{\pm 7.09}$
& ${84.35}_{\pm 0.72}$ & ${88.56}_{\pm 0.21}$ \\

SSIAT~\cite{ssiat} & CVPR'24 & \cmark
& ${80.52}_{\pm 0.07}$ & ${84.25}_{\pm 0.31}$ 
& ${75.67}_{\pm 0.14}$ & ${82.30}_{\pm 0.36}$ 
& ${59.16}_{\pm 1.03}$ & $\underline{68.45}_{\pm 1.92}$
& ${85.11}_{\pm 0.56}$ & ${89.80}_{\pm 0.34}$ \\

MOS~\cite{mos} & AAAI'25 & \cmark
& ${78.76}_{\pm 0.17}$ & ${82.37}_{\pm 0.26}$
& ${75.16}_{\pm 0.59}$ & ${80.53}_{\pm 0.70}$
& -- & --
& -- & -- \\

MACIL~\cite{macil} & ICML'25 & \cmark 
& $\mathbf{83.37}_{\bm{\pm}\mathbf{0.26}}$ & $\underline{87.00}_{\pm 0.35}$ 
& $\underline{79.43}_{\pm 0.34}$ & $\underline{84.34}_{\pm 0.32}$ 
& $\underline{59.36}_{\pm 1.27}$ & ${68.03}_{\pm 2.00}$
& $\underline{86.62}_{\pm 0.35}$ & $\underline{90.88}_{\pm 0.20}$ \\

\midrule
$\textbf{$\textbf{M}^2$LE (Ours) + CA}$ & -- & \cmark 
& $\underline{82.74}_{\pm 0.29}$ & $\mathbf{87.25}_{\bm{\pm}\mathbf{0.28}}$
& $\mathbf{81.12}_{\bm{\pm}\mathbf{0.14}}$ & $\mathbf{86.12}_{\bm{\pm}\mathbf{0.33}}$
& $\mathbf{64.25}_{\bm{\pm}\mathbf{0.56}}$ & $\mathbf{72.54}_{\bm{\pm}\mathbf{0.92}}$
& $\mathbf{87.61}_{\bm{\pm}\mathbf{0.13}}$ & $\mathbf{91.51}_{\bm{\pm}\mathbf{0.18}}$ \\
\bottomrule
\end{tabular}
}
\end{table*}

\textbf{Datasets.}
We evaluate our method on five standard continual learning benchmarks: ImageNet-R~\cite{-r}, ImageNet-A~\cite{-a}, CIFAR100~\cite{cifar100}, CUB200~\cite{cub}, and DomainNet~\cite{domainnet}. 
ImageNet-R contains 30,000 images from 200 classes with diverse styles, while ImageNet-A consists of 7,500 natural adversarial examples spanning 200 classes. 
CIFAR100 is a standard benchmark with 60,000 images from 100 classes, and CUB200 is a fine-grained bird classification dataset with 200 categories. 
For DomainNet, following prior works~\cite{loda2026}, we use the commonly adopted 200-class evaluation protocol.

\textbf{Evaluation Metrics.}
Following prior PEFT-based continual learning works~\cite{macil,ssiat}, we adopt two standard evaluation metrics: (1) $\mathcal{A}_{{Last}}$, i.e., the final accuracy over all classes after the last task, and (2) $\mathcal{A}_{{Avg}}$, i.e., the average accuracy across all incremental tasks.

\textbf{Implementation Details.}
We use ViT-B/16~\cite{ptm} pre-trained on ImageNet-21K~\cite{imagenet-21k} as the backbone. All experiments are conducted on a single NVIDIA RTX 3090 GPU. 
The model is optimized with SGD and a Cosine Annealing scheduler following~\cite{macil}. 
We train each task for 20 epochs with a batch size of 48. The initial learning rate is set to 0.02 on ImageNet-R and 0.01 on the other datasets. 
For the HIM, the hyper-parameter $\lambda$ in Eq.~\eqref{eq:element_wise_solution} is set to 5 for all datasets. 
For the TSP, $\rho$ in Eq.~\eqref{eq:meta_probe} is set to 0.3 on ImageNet-A and 0.5 on the other datasets, while $\beta$ in Eq.~\eqref{eq:meta_mix} is set to 0.6 on ImageNet-A and 0.8 on the other datasets.

\textbf{Variant with classifier alignment.}
For fair comparison with methods that use feature replay or classifier alignment (CA), we also evaluate an enhanced variant, denoted as $\text{M}^2$LE+CA.
$\text{M}^2$LE follows the end-to-end continual learning protocol without feature replay or post-hoc classifier adjustment.
$\text{M}^2$LE+CA further applies the widely used classifier alignment strategy after each task, following SLCA~\cite{slca}.
We report these two variants separately to distinguish the effect of our proposed method from the additional classifier alignment stage.

\subsection{Comparison with the State-of-the-Arts}
We compare our method with representative state-of-the-art continual learning approaches from three categories, including prompt-based methods~\cite{l2p,codaprompt}, adapter-based methods~\cite{ssiat,saef,sema,mos} and LoRA-based methods~\cite{inflora,plan,sdlora,bilora,pm,ewc-lora,macil}. 
For completeness, methods that rely on feature replay or classifier alignment are reported separately from strictly rehearsal-free methods.

As shown in Tables~\ref{tab:main_10s} and \ref{tab:main_table_2}, our method demonstrates clear advantages over existing PEFT-based continual learning methods. 
Without classifier alignment, compared with the LoRA-P\&M, $\text{M}^2$LE consistently improves $\mathcal{A}_{{Last}}$ across all settings, with gains of $1.39\%$--$5.88\%$ on the 10 task benchmarks and $0.43\%$--$2.25\%$ under the additional task settings. 
With classifier alignment, $\text{M}^2$LE+CA achieves highly competitive performance against feature-replay-based methods, outperforming MACIL in $\mathcal{A}_{{Last}}$ by $0.80\%$--$1.82\%$ on the 10 task ImageNet-R/ImageNet-A settings and by $1.69\%$--$4.89\%$ on their 20 task settings.
The margins are relatively smaller on CIFAR100 and CUB, where strong pre-trained representations already provide discriminative features and thus reduce the need for further backbone adaptation.

\begin{table*}[t]
\centering
\caption{Ablation studies on ImageNet-R and ImageNet-A under 10 and 20 incremental tasks.}
\label{tab:ablation_main}
\scriptsize
\setlength{\tabcolsep}{3pt}
\resizebox{\textwidth}{!}{%
\begin{tabular}{c c c cc cc cc cc}
\toprule
\multirow{2}{*}{Idx} & \multirow{2}{*}{HIM} & \multirow{2}{*}{TSP}
& \multicolumn{2}{c}{10S-ImageNet-R}
& \multicolumn{2}{c}{10S-ImageNet-A}
& \multicolumn{2}{c}{20S-ImageNet-R}
& \multicolumn{2}{c}{20S-ImageNet-A} \\
& &
& $\mathcal{A}_{Last}\uparrow$ & $\mathcal{A}_{Avg}\uparrow$
& $\mathcal{A}_{Last}\uparrow$ & $\mathcal{A}_{Avg}\uparrow$
& $\mathcal{A}_{Last}\uparrow$ & $\mathcal{A}_{Avg}\uparrow$
& $\mathcal{A}_{Last}\uparrow$ & $\mathcal{A}_{Avg}\uparrow$ \\
\midrule
1 & -- & --
& 73.05$_{ \pm 0.35}$ & 80.40$_{ \pm 0.60}$
& 52.56$_{ \pm 2.08}$ & 63.36$_{ \pm 1.32}$
& 66.14$_{ \pm 0.20}$ & 73.72$_{ \pm 0.25}$
& 46.06$_{ \pm 2.14}$ & 59.43$_{ \pm 1.77}$ \\

2 & \checkmark & --
& 80.36$_{ \pm 0.40}$ & 85.78$_{ \pm 0.44}$
& 61.35$_{ \pm 1.06}$ & 70.23$_{ \pm 2.04}$
& 78.04$_{ \pm 0.35}$ & 84.31$_{ \pm 0.54}$
& 52.80$_{ \pm 1.46}$ & 62.64$_{ \pm 2.14}$ \\

3 & -- & \checkmark
& 74.20$_{ \pm 0.91}$ & 80.67$_{ \pm 0.47}$
& 56.57$_{ \pm 2.31}$ & 66.26$_{ \pm 1.46}$
& 69.17$_{ \pm 0.10}$ & 77.16$_{ \pm 0.39}$
& 49.50$_{ \pm 0.72}$ & 60.45$_{ \pm 1.52}$ \\

4 & \checkmark & \checkmark
& $\mathbf{81.34}_{ \bm{\pm}\mathbf{0.16}}$ & $\mathbf{86.53}_{ \bm{\pm}\mathbf{0.51}}$
& $\mathbf{62.45}_{ \bm{\pm}\mathbf{1.34}}$ & $\mathbf{71.02}_{ \bm{\pm}\mathbf{1.82}}$
& $\mathbf{78.62}_{ \bm{\pm}\mathbf{0.43}}$ & $\mathbf{84.62}_{ \bm{\pm}\mathbf{0.26}}$
& $\mathbf{53.87}_{ \bm{\pm}\mathbf{1.45}}$ & $\mathbf{63.69}_{ \bm{\pm}\mathbf{1.87}}$ \\
\bottomrule
\end{tabular}%
}
\end{table*}
\begin{table}[t]
\centering
\caption{Comparison of different merge strategies on 10S-ImageNet-R and 10S-ImageNet-A.}
\label{tab:merge_strategy}
\resizebox{\linewidth}{!}{
\begin{tabular}{l|cc|cc}
\toprule
\multirow{2}{*}{Strategy} 
& \multicolumn{2}{c|}{10S-ImageNet-R}
& \multicolumn{2}{c}{10S-ImageNet-A} \\
& $\mathcal{A}_{Last}\uparrow$ & $\mathcal{A}_{Avg}\uparrow$
& $\mathcal{A}_{Last}\uparrow$ & $\mathcal{A}_{Avg}\uparrow$ \\
\midrule
Seq-FT
& $73.05_{\pm0.35}$ & $80.40_{\pm0.60}$ 
& $52.56_{\pm2.08}$ & $63.36_{\pm1.32}$ \\
EMA-style ($\theta=\frac{t-1}{t}\theta_{t-1}+\frac{1}{t}\theta_{t}$)~\cite{ema}
& $79.14_{\pm0.10}$ & $85.08_{\pm0.41}$
& $59.64_{\pm1.22}$ & $69.52_{\pm1.82}$ \\
LoRA-P\&M~\cite{pm}
& $78.89_{\pm0.44}$ & $84.99_{\pm0.45}$
& $59.93_{\pm1.57}$ & $69.87_{\pm2.05}$ \\
HIM-last 
($\mathcal{J}_{old}=\mathcal{L}_{t-1}(\mathbf{W}_{t-1}+\hat{\Delta\mathbf{W}}_t)-\mathcal{L}_{t-1}(\mathbf{W}_{t-1})$)
& $76.23_{\pm0.76}$ & $83.81_{\pm0.61}$
& $58.17_{\pm2.38}$ & $68.84_{\pm1.93}$ \\
\textbf{HIM (Ours)}
& $\mathbf{80.36}_{ \bm{\pm}\mathbf{0.40}}$ & $\mathbf{85.78}_{ \bm{\pm}\mathbf{0.44}}$
& $\mathbf{61.35}_{ \bm{\pm}\mathbf{1.06}}$ & $\mathbf{70.23}_{ \bm{\pm}\mathbf{2.04}}$ \\
\bottomrule
\end{tabular}
}
\end{table}


\begin{table}[t]
\centering
\caption{Comparison of different perturbation strategies on 10S-ImageNet-R and 10S-ImageNet-A.}
\label{tab:meta_strategy}
\begin{adjustbox}{max width=0.6\textwidth}
\begin{tabular}{l|cc|cc}
\toprule
\multirow{2}{*}{Strategy} 
& \multicolumn{2}{c|}{10S-ImageNet-R}
& \multicolumn{2}{c}{10S-ImageNet-A} \\
& $\mathcal{A}_{Last}\uparrow$ & $\mathcal{A}_{Avg}\uparrow$
& $\mathcal{A}_{Last}\uparrow$ & $\mathcal{A}_{Avg}\uparrow$ \\
\midrule
w/o TSP
& $80.36_{\pm0.40}$ & $85.78_{\pm0.44}$ 
& $61.35_{\pm1.06}$ & $70.23_{\pm2.04}$ \\
SAM~\cite{sam}
& $79.55_{\pm0.17}$ & $84.92_{\pm0.44}$
& $60.46_{\pm1.51}$ & $69.00_{\pm1.98}$ \\
\textbf{TSP (Ours)}
& $\mathbf{81.34}_{ \bm{\pm}\mathbf{0.16}}$ & $\mathbf{86.53}_{ \bm{\pm}\mathbf{0.51}}$
& $\mathbf{62.45}_{ \bm{\pm}\mathbf{1.34}}$ & $\mathbf{71.02}_{ \bm{\pm}\mathbf{1.82}}$ \\
\bottomrule

\end{tabular}
\end{adjustbox}
\end{table}

\subsection{Ablation Studies}

We conduct ablation studies on the main components of our training pipeline, including the sequential fine-tuning baseline, the proposed HIM, and TSP.
As shown in Tab.~\ref{tab:ablation_main}, HIM (Idx 2) brings the main improvements over the baseline (Idx 1), increasing $\mathcal{A}_{Last}$ by 6.74\%--11.90\% across all four settings. 
This shows that reweighting the current task update under the joint loss impact objective in Eq.~\eqref{eq:ideal_obj} is more effective than directly adding it to the previous inference model. 
TSP (Idx 3) also consistently improves the baseline, with $\mathcal{A}_{{Last}}$ gains of 1.15\%--4.01\%. 
The improvement is particularly clear on the more challenging ImageNet-A benchmark, suggesting that TSP helps the current task update remain stable under task simulating perturbations, making it less sensitive to subsequent merge-induced changes.
Notably, combining HIM and TSP (Idx 4) further improves $\mathcal{A}_{{Last}}$ by 0.58\%--1.10\% over HIM alone and achieves the best overall performance across all settings, indicating that the two components play complementary roles.

\begin{table*}[t]
\centering
\caption{Task-by-task results (\%) on ImageNet-R and ImageNet-A under the 10 task setting. We compare the results with and without TSP. The forgetting metric is defined in Appendix~\ref{forget}}
\label{tab:meta_task_imagenetr_a}
\resizebox{\textwidth}{!}{
\begin{tabular}{c|c|c|cccccccccc}
\toprule
Dataset & Metrics & Methods 
& Task 0 & Task 1 & Task 2 & Task 3 & Task 4 
& Task 5 & Task 6 & Task 7 & Task 8 & Task 9 \\
\midrule
\multirow{6}{*}{ImageNet-R}
& \multirow{3}{*}{Accuracy $\uparrow$}
& w/o TSP
& $94.96_{\pm1.23}$ & $91.31_{\pm1.22}$ & $89.48_{\pm0.20}$ & $86.91_{\pm0.67}$ & $84.73_{\pm0.64}$
& $83.68_{\pm0.20}$ & $83.00_{\pm0.61}$ & $82.13_{\pm0.53}$ & $81.24_{\pm0.56}$ & $80.36_{\pm0.40}$ \\
&& w/ TSP
& $\mathbf{95.14}_{\pm1.07}$ & $\mathbf{92.14}_{\pm0.86}$ & $\mathbf{89.93}_{\pm0.85}$ & $\mathbf{87.58}_{\pm0.56}$ & $\mathbf{85.86}_{\pm0.91}$
& $\mathbf{84.49}_{\pm0.61}$ & $\mathbf{83.67}_{\pm0.42}$ & $\mathbf{82.98}_{\pm0.40}$ & $\mathbf{82.14}_{\pm0.54}$ & $\mathbf{81.34}_{\pm0.16}$ \\
&& \textcolor{ForestGreen}{\textbf{$\Delta$}}
& \textcolor{ForestGreen}{\textbf{$+0.18$}} & \textcolor{ForestGreen}{\textbf{$+0.83$}} & \textcolor{ForestGreen}{\textbf{$+0.45$}} & \textcolor{ForestGreen}{\textbf{$+0.67$}} & \textcolor{ForestGreen}{\textbf{$+1.13$}}
& \textcolor{ForestGreen}{\textbf{$+0.81$}} & \textcolor{ForestGreen}{\textbf{$+0.67$}} & \textcolor{ForestGreen}{\textbf{$+0.85$}} & \textcolor{ForestGreen}{\textbf{$+0.90$}} & \textcolor{ForestGreen}{\textbf{$+0.98$}} \\
\cmidrule(lr){2-13}
& \multirow{3}{*}{Forgetting $\downarrow$}
& w/o TSP
& -- & $4.98_{\pm0.90}$ & $4.87_{\pm0.60}$ & $6.16_{\pm0.58}$ & $7.00_{\pm0.59}$
& $7.34_{\pm0.20}$ & $7.26_{\pm0.67}$ & $7.34_{\pm0.49}$ & $7.43_{\pm0.44}$ & $7.76_{\pm0.38}$ \\
&& w/ TSP
& -- & $\mathbf{3.17}_{\pm0.60}$ & $\mathbf{3.79}_{\pm0.86}$ & $\mathbf{4.40}_{\pm0.58}$ & $\mathbf{4.89}_{\pm0.78}$
& $\mathbf{5.41}_{\pm0.24}$ & $\mathbf{5.46}_{\pm0.62}$ & $\mathbf{5.49}_{\pm0.88}$ & $\mathbf{5.52}_{\pm0.82}$ & $\mathbf{5.77}_{\pm0.64}$ \\
&& \textcolor{Orange}{\textbf{$\Delta$}}
& -- & \textcolor{Orange}{\textbf{$-1.81$}} & \textcolor{Orange}{\textbf{$-1.08$}} & \textcolor{Orange}{\textbf{$-1.76$}} & \textcolor{Orange}{\textbf{$-2.11$}}
& \textcolor{Orange}{\textbf{$-1.93$}} & \textcolor{Orange}{\textbf{$-1.80$}} & \textcolor{Orange}{\textbf{$-1.85$}} & \textcolor{Orange}{\textbf{$-1.91$}} & \textcolor{Orange}{\textbf{$-1.99$}} \\
\midrule
\multirow{6}{*}{ImageNet-A}
& \multirow{3}{*}{Accuracy $\uparrow$}
& w/o TSP
& $86.56_{\pm1.34}$ & $80.06_{\pm1.70}$ & $74.96_{\pm1.79}$ & $71.95_{\pm3.44}$ & $69.65_{\pm3.18}$
& $67.65_{\pm2.97}$ & $64.79_{\pm2.13}$ & $63.60_{\pm1.98}$ & $61.75_{\pm1.88}$ & $61.35_{\pm1.06}$ \\
&& w/ TSP
& $\mathbf{86.70}_{\pm0.30}$ & $\mathbf{81.60}_{\pm0.54}$ & $\mathbf{75.84}_{\pm2.67}$ & $\mathbf{72.28}_{\pm2.89}$ & $\mathbf{70.06}_{\pm2.83}$
& $\mathbf{68.60}_{\pm2.70}$ & $\mathbf{65.83}_{\pm2.63}$ & $\mathbf{64.07}_{\pm1.58}$ & $\mathbf{62.79}_{\pm1.64}$ & $\mathbf{62.45}_{\pm1.34}$ \\
&& \textcolor{ForestGreen}{\textbf{$\Delta$}}
& \textcolor{ForestGreen}{\textbf{$+0.14$}} & \textcolor{ForestGreen}{\textbf{$+1.54$}} & \textcolor{ForestGreen}{\textbf{$+0.88$}} & \textcolor{ForestGreen}{\textbf{$+0.33$}} & \textcolor{ForestGreen}{\textbf{$+0.41$}}
& \textcolor{ForestGreen}{\textbf{$+0.95$}} & \textcolor{ForestGreen}{\textbf{$+1.04$}} & \textcolor{ForestGreen}{\textbf{$+0.47$}} & \textcolor{ForestGreen}{\textbf{$+1.04$}} & \textcolor{ForestGreen}{\textbf{$+1.10$}} \\
\cmidrule(lr){2-13}
& \multirow{3}{*}{Forgetting $\downarrow$}
& w/o TSP
& -- & $13.54_{\pm6.63}$ & $11.43_{\pm1.91}$ & $12.49_{\pm1.06}$ & $12.71_{\pm0.85}$
& $12.27_{\pm0.15}$ & $13.88_{\pm0.72}$ & $13.47_{\pm0.44}$ & $14.19_{\pm1.52}$ & $14.01_{\pm0.90}$ \\
&& w/ TSP
& -- & $\mathbf{10.56}_{\pm0.60}$ & $\mathbf{9.19}_{\pm2.12}$ & $\mathbf{9.98}_{\pm1.49}$ & $\mathbf{9.95}_{\pm1.04}$
& $\mathbf{9.24}_{\pm0.91}$ & $\mathbf{10.52}_{\pm1.26}$ & $\mathbf{10.77}_{\pm0.69}$ & $\mathbf{10.49}_{\pm0.86}$ & $\mathbf{10.04}_{\pm1.29}$ \\
&& \textcolor{Orange}{\textbf{$\Delta$}}
& -- & \textcolor{Orange}{\textbf{$-2.98$}} & \textcolor{Orange}{\textbf{$-2.24$}} & \textcolor{Orange}{\textbf{$-2.51$}} & \textcolor{Orange}{\textbf{$-2.76$}}
& \textcolor{Orange}{\textbf{$-3.03$}} & \textcolor{Orange}{\textbf{$-3.36$}} & \textcolor{Orange}{\textbf{$-2.70$}} & \textcolor{Orange}{\textbf{$-3.70$}} & \textcolor{Orange}{\textbf{$-3.97$}} \\
\bottomrule
\end{tabular}
}
\end{table*}
\begin{figure*}[htbp]
\centering
\vspace{-3.0mm}
\includegraphics[width=1\textwidth]{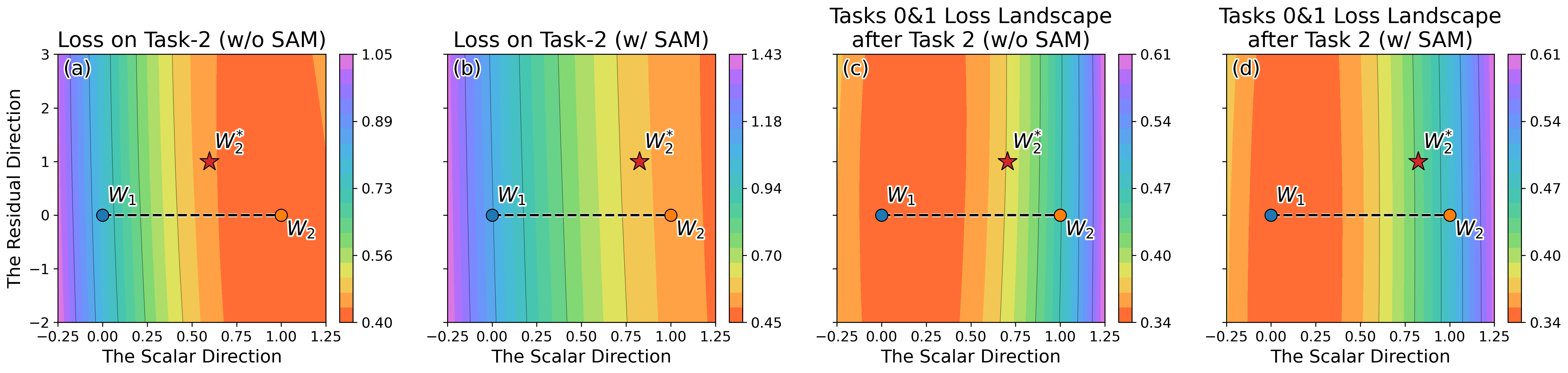}
\caption{Analysis of the effect of SAM on 10S-ImageNet-R.
Panels (a,b) visualize the task 2 loss landscapes for updates learned without/with SAM, while panels (c,d) show the total old task loss after merging. 
The dashed line denotes scalar interpolation, and the red star marks the merged solution.
}
\label{fig:analysis_sam}
\vspace{-4.0mm}
\end{figure*}

\begin{table*}[h]
\centering
\caption{{Computational and memory overhead on 10S-ImageNet-A.}}
\label{tab:efficiency_tsp}
\begin{adjustbox}{max width=1\textwidth}
\begin{tabular}{lcccc}
\toprule
Methods 
& \makecell{GPU Memory (MiB)}
& \makecell{Time/Batch (s)}
& \makecell{Extra Memory (MB)}
& $\mathcal{A}_{Last}$ $\uparrow$ \\
\midrule
Seq-FT 
& 4560 & 0.12 & 0.0 
& $52.56_{\pm 2.08}$ \\

TSP 
& 2881 & 0.34 & 0.0 
& $56.57_{\pm 2.31}$ \\

HIM 
& 4560 & 0.12 & 54.0 
& $61.35_{\pm 1.06}$ \\

HIM w/ SAM 
& 4568 & 0.26 & 54.0 
& $60.46_{\pm 1.51}$ \\

\textbf{HIM w/ TSP} 
& \textbf{2881} & 0.34 & 54.0 
& $\mathbf{62.45_{\pm 1.34}}$ \\

\bottomrule
\end{tabular}
\end{adjustbox}
\end{table*}

\subsection{In-Depth Analysis}

\label{in_depth_analysis}

\textbf{Comparison of different merge strategies.}
We further compare different merge strategies on 10S-ImageNet-R and 10S-ImageNet-A to analyze how the choice of merge rule affects continual learning performance.
Seq-FT is used as a without merge baseline that updates the inference model by directly adding the current task LoRA update without any merge rule.
As shown in Tab.~\ref{tab:merge_strategy}, EMA-style and LoRA-P\&M both outperform Seq-FT, confirming the benefit of merging in rehearsal-free CL.

However, these methods still rely on a scalar factor shared by all parameters, matching the limitation discussed in Sec.~\ref{pre_analysis}.
In contrast, HIM directly solves the target merged update under Eq.~\eqref{eq:ideal_obj} and obtains the element-wise solution in Eq.~\eqref{eq:element_wise_solution}. 
It achieves the best results, improving $\mathcal{A}_{Last}$ over EMA-style averaging by 1.22\%/1.71\% and over LoRA-P\&M by 1.47\%/1.42\% on ImageNet-R/ImageNet-A, showing the assigning merge strengths at the element level is more effective than using a shared scalar factor.
We further compare HIM with a HIM variant (HIM-last) that only uses task $t-1$ in Eq.~\eqref{eq:old_task_term}; without explicit constraints from earlier tasks, this variant suffers from more severe forgetting, indicating that accumulated old task information in $\mathbf{F}_{1:t-1}$ is important for merging.
More HIM variants are reported in Appendix~\ref{more_old}.

\textbf{Analysis of perturbation training under cross-task merging.}
We further compare TSP with SAM~\cite{sam} and without TSP to examine the effect of different perturbation training strategies during subsequent merging.
As shown in Tab.~\ref{tab:meta_strategy}, SAM performs worse than HIM alone. 
Compared with HIM alone, TSP improves $\mathcal{A}_{Last}$ by $0.98\%$/$1.10\%$ on ImageNet-R/ImageNet-A.
To better understand this result, we visualize its behavior on both the current task and the old tasks after merging.
Fig.~\ref{fig:analysis_sam} shows that SAM can still obtain a low loss solution on task 2, but the merged solution falls into a less favorable region for old tasks 0 and 1.
This indicates that optimizing current task sharpness cannot well capture the cross-task interference caused by sequential LoRA updates.
In contrast, TSP constructs the perturbation from one label-disjoint subset and evaluates it on the other subset.
As a result, our method achieves the best performance, improving $\mathcal{A}_{Last}$ over SAM by $1.79\%$/$1.99\%$ on ImageNet-R/ImageNet-A.

\textbf{Effect of TSP on  forgetting.}
TSP aims to reduce the sensitivity of the current task update to the parameter change introduced by the subsequent merge.
As shown in Tab.~\ref{tab:meta_task_imagenetr_a}, adding TSP consistently improves total accuracy and reduces forgetting across incremental stages on both ImageNet-R and ImageNet-A.
The final forgetting decreases from $7.76\%$ to $5.77\%$ on ImageNet-R and from $14.01\%$ to $10.04\%$ on ImageNet-A.
This supports our motivation that stabilizing the current task update can alleviate forgetting during continual merging.
More hyper-parameter analyses and results with different backbones are provided in Appendix~\ref{hyper} and Appendix~\ref{clip}.

\textbf{Computational and Memory Overhead.}
Table~\ref{tab:efficiency_tsp} compares the overhead of HIM and TSP on 10S-ImageNet-A.
HIM requires 54.0 MB of additional Fisher storage, independent of the number of tasks, and 3.8 s per task for Fisher estimation.
TSP increases the training time per batch from 0.12 s to 0.34 s, but reduces GPU memory usage from 4560 MiB to 2881 MiB.
These costs are limited to training, with no additional inference overhead.
Further results on the effect of LoRA on overhead and accuracy are provided in Appendix~\ref{rank}.

\subsection{Further Analysis}

\label{analysis}

\begin{figure*}[htbp]
\centering
\includegraphics[width=1\textwidth]{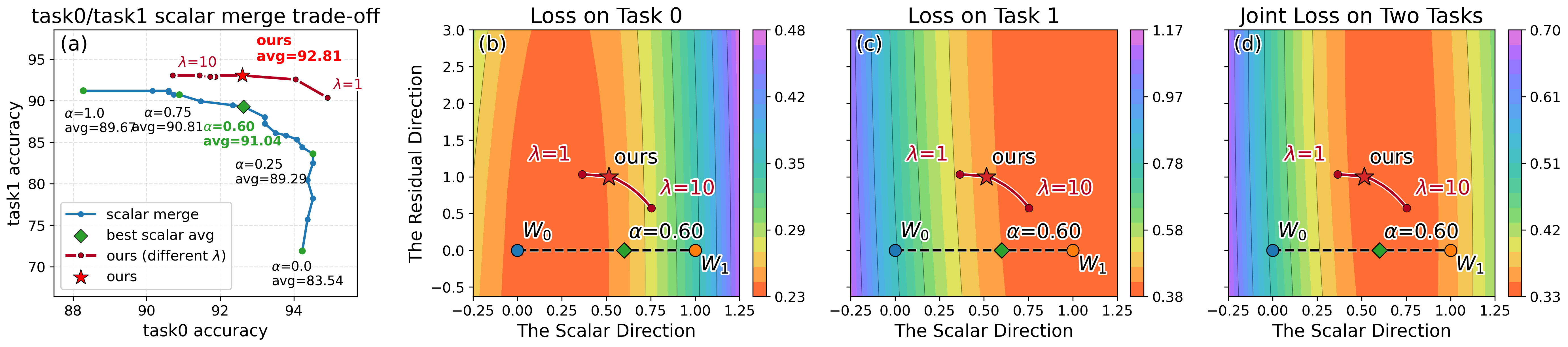}
\caption{Analysis of scalar merge and HIM without TSP on 10S-ImageNet-R.
Panel (a) compares task 0/1 accuracy under different merge choices, and panels (b--d) show the task 0, task 1, and joint loss landscapes.
The dashed line denotes scalar interpolation; the green diamond and red star mark the best scalar merge and selected HIM solution, respectively.
}
\vspace{-2.0mm}
\label{fig:analysis_merge}
\end{figure*}

\begin{figure*}[htbp]
\centering
\vspace{-1.0mm}
\includegraphics[width=1\textwidth]{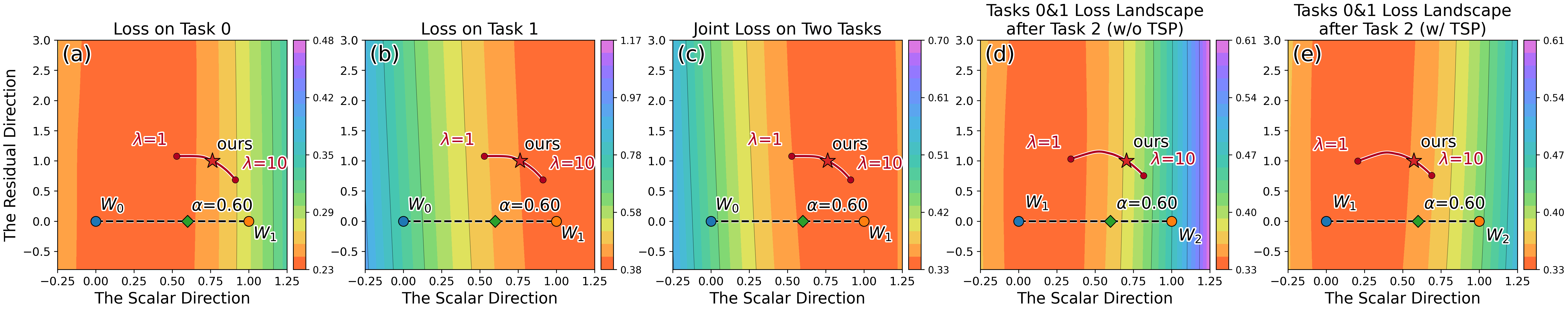}
\caption{Analysis of the effect of TSP on 10S-ImageNet-R. 
Panels (a--c) show the task 0, task 1, and joint loss landscapes after introducing TSP, corresponding to Fig.~\ref{fig:analysis_merge}(b--d).
Panels (d,e) isolate TSP at the task 2 merge step: from the same task 1 model, we learn and merge the task 2 update without/with TSP and visualize the joint loss over old tasks 0 and 1.
The dashed line denotes scalar interpolation, and the red star marks the merged solution.
}
\label{fig:analysis_meta}
\vspace{-5.0mm}
\end{figure*}

\textbf{Further analysis of HIM and TSP.}
We further analyze the behaviors of HIM and TSP using the task 1 accuracy against task 0 accuracy and the corresponding loss landscape visualizations.
For the loss landscapes, we construct a 2D plane to compare the scalar interpolation path with our merged solution.
Let $\mathbf{W}_1$ denote the model obtained by directly adding the task 1 update to $\mathbf{W}_0$, and define the scalar interpolation direction as $\mathbf{u} = \mathbf{W}_1 - \mathbf{W}_0$.
Let $\mathbf{d} = \mathbf{W}_{\mathrm{ours}} - \mathbf{W}_0$ denote the displacement from $\mathbf{W}_0$ to our selected merged solution.
We then remove from $\mathbf{d}$ its component along $\mathbf{u}$ and obtain the residual direction
$\mathbf{r} = \mathbf{d} - \frac{\langle \mathbf{d},\mathbf{u}\rangle}{\langle \mathbf{u},\mathbf{u}\rangle}\mathbf{u}$ and $\mathbf{v} = \mathbf{r}/\|\mathbf{r}\|_2$, where $\mathbf{v}$ is orthogonal to the scalar interpolation direction.
The loss is visualized on the plane $\mathbf{W}(x,y)=\mathbf{W}_0+x\mathbf{u}+y\mathbf{v}$.

\textbf{Observation 1: HIM reaches a better solution beyond scalar interpolation.}
Fig.~\ref{fig:analysis_merge}(a) shows that scalar merge is restricted to a interpolation path, while varying $\lambda$ in our method produces solutions outside this path.
This indicates that HIM provides a more flexible search space than a shared scalar factor.
More importantly, the joint loss landscape in Fig.~\ref{fig:analysis_merge}(d) shows that the red curve produced by varying $\lambda$ mostly remains in the low loss region, rather than being restricted to the scalar path.
This suggests that the merged update under the joint objective can provide more flexible balances between previous task and current task.

\textbf{Observation 2: TSP improves the local stability of the merged solution.}
Compared with Fig.~\ref{fig:analysis_merge}(b)--(d), Fig.~\ref{fig:analysis_meta}(a)--(c) show that introducing TSP makes the low loss regions around the merged solution visibly wider for task 0, task 1, and the joint loss.
This suggests that TSP makes the learned update less sensitive to local parameter changes before merging.
To further isolate the effect of TSP in the next merge step, Fig.~\ref{fig:analysis_meta}(d) and Fig.~\ref{fig:analysis_meta}(e) start from the same model before learning task 2 and visualize the joint loss over old tasks 0 and 1 after merging the task 2 update without and with TSP, respectively.
With TSP, the old task low loss region becomes wider, and the merged solution is located within more stable region, indicating better robustness to the parameter change introduced by the task 2 merge.
\section{Conclusion}
In this work, we show that different parameters in  LoRA update may prefer different merge strengths.
Motivated by this, we propose $\text{M}^2$LE, which directly solves for the target merged update, yielding an analytic element-wise modulation of the current LoRA update. $\text{M}^2$LE first derives Hessian Information-guided Adaptive Merging (HIM), an analytic element-wise modulation rule under a joint loss objective over old and new tasks. 
It further introduces Task Simulation Perturbation (TSP) to simulate cross-task interference during training and improve the stability during subsequent merging. 
Overall, $\text{M}^2$LE offers a new view for continual learning with LoRA, showing that adapting merge strengths for different LoRA elements provides an effective way to balance stability and plasticity.
\begin{ack}
This work was supported in part by the National Key R\&D Program of China under Grant No.2023YFA1008600, in part by the National Natural Science Foundation of China under Grants 62576262, in part by the Natural Science Foundation Basic Research Program of Shaanxi Province under grant 2026JC-YXQN-202, in part by Scientific and Technological Innovation Teams in Shaanxi Province under Grant 2025RS-CXTD-011, in part by the Shaanxi Province Core Technology Research and Development Project under Grant 2024QY2-GJHX-11, in part by the Natural Science Basic Research Program of Shaanxi under Grant 2026SYS-SYSZD-105, in part by the Key Research and Development Program of Shaanxi under Grant 2026JC-ZDGG-42, in part by the Establishment Fund of the State Key Laboratory of Wakefulness/Sleep and Cognition (Chongqing Institute for Brain and Intelligence), in part by the Fundamental Research Funds for the Central Universities under Grant QTZX25083 and QTZX26138.
\end{ack}

\bibliographystyle{plainnat}
\bibliography{neurips_2026}

@String(ICLR = {Int. Conf. Learn. Represent.})

@String(AAAI = {AAAI})

@String(ICLR  = {ICLR})

@inproceedings{ssiat,
  title={Semantically-shifted incremental adapter-tuning is a continual vitransformer},
  author={Tan, Yuwen and Zhou, Qinhao and Xiang, Xiang and Wang, Ke and Wu, Yuchuan and Li, Yongbin},
  booktitle={Proceedings of the IEEE/CVF Conference on Computer Vision and Pattern Recognition},
  pages={23252--23262},
  year={2024}
}

@article{pm,
  title={Train with Perturbation, Infer after Merging: A Two-Stage Framework for Continual Learning},
  author={Qiu, Haomiao and Zhang, Miao and Qiao, Ziyue and Nie, Liqiang},
  journal={arXiv preprint arXiv:2505.22389},
  year={2025}
}

@article{macil,
  title={Navigating semantic drift in task-agnostic class-incremental learning},
  author={Wu, Fangwen and Cheng, Lechao and Tang, Shengeng and Zhu, Xiaofeng and Fang, Chaowei and Zhang, Dingwen and Wang, Meng},
  journal={arXiv preprint arXiv:2502.07560},
  year={2025}
}

@article{sdlora,
  title={Sd-lora: Scalable decoupled low-rank adaptation for class incremental learning},
  author={Wu, Yichen and Piao, Hongming and Huang, Long-Kai and Wang, Renzhen and Li, Wanhua and Pfister, Hanspeter and Meng, Deyu and Ma, Kede and Wei, Ying},
  journal={arXiv preprint arXiv:2501.13198},
  year={2025}
}

@inproceedings{sema,
  title={Self-expansion of pre-trained models with mixture of adapters for continual learning},
  author={Wang, Huiyi and Lu, Haodong and Yao, Lina and Gong, Dong},
  booktitle={Proceedings of the Computer Vision and Pattern Recognition Conference},
  pages={10087--10098},
  year={2025}
}

@inproceedings{ema,
  title={Towards better plasticity-stability trade-off in incremental learning: A simple linear connector},
  author={Lin, Guoliang and Chu, Hanlu and Lai, Hanjiang},
  booktitle={Proceedings of the IEEE/CVF conference on computer vision and pattern recognition},
  pages={89--98},
  year={2022}
}

@inproceedings{plan,
  title={PLAN: Proactive Low-Rank Allocation for Continual Learning},
  author={Wang, Xiequn and Zhuang, Zhan and Zhang, Yu},
  booktitle={Proceedings of the IEEE/CVF International Conference on Computer Vision},
  pages={2909--2918},
  year={2025}
}

@inproceedings{dualprompt,
  title={Dualprompt: Complementary prompting for rehearsal-free continual learning},
  author={Wang, Zifeng and Zhang, Zizhao and Ebrahimi, Sayna and Sun, Ruoxi and Zhang, Han and Lee, Chen-Yu and Ren, Xiaoqi and Su, Guolong and Perot, Vincent and Dy, Jennifer and others},
  booktitle={European conference on computer vision},
  pages={631--648},
  year={2022},
  organization={Springer}
}

@inproceedings{l2p,
  title={Learning to prompt for continual learning},
  author={Wang, Zifeng and Zhang, Zizhao and Lee, Chen-Yu and Zhang, Han and Sun, Ruoxi and Ren, Xiaoqi and Su, Guolong and Perot, Vincent and Dy, Jennifer and Pfister, Tomas},
  booktitle={Proceedings of the IEEE/CVF conference on computer vision and pattern recognition},
  pages={139--149},
  year={2022}
}

@inproceedings{codaprompt,
  title={Coda-prompt: Continual decomposed attention-based prompting for rehearsal-free continual learning},
  author={Smith, James Seale and Karlinsky, Leonid and Gutta, Vyshnavi and Cascante-Bonilla, Paola and Kim, Donghyun and Arbelle, Assaf and Panda, Rameswar and Feris, Rogerio and Kira, Zsolt},
  booktitle={Proceedings of the IEEE/CVF conference on computer vision and pattern recognition},
  pages={11909--11919},
  year={2023}
}

@inproceedings{inflora,
  title={Inflora: Interference-free low-rank adaptation for continual learning},
  author={Liang, Yan-Shuo and Li, Wu-Jun},
  booktitle={Proceedings of the IEEE/CVF Conference on Computer Vision and Pattern Recognition},
  pages={23638--23647},
  year={2024}
}

@inproceedings{slca,
  title={Slca: Slow learner with classifier alignment for continual learning on a pre-trained model},
  author={Zhang, Gengwei and Wang, Liyuan and Kang, Guoliang and Chen, Ling and Wei, Yunchao},
  booktitle={Proceedings of the IEEE/CVF International Conference on Computer Vision},
  pages={19148--19158},
  year={2023}
}

@inproceedings{mos,
  title={Mos: Model surgery for pre-trained model-based class-incremental learning},
  author={Sun, Hai-Long and Zhou, Da-Wei and Zhao, Hanbin and Gan, Le and Zhan, De-Chuan and Ye, Han-Jia},
  booktitle={Proceedings of the AAAI Conference on Artificial Intelligence},
  volume={39},
  number={19},
  pages={20699--20707},
  year={2025}
}

@article{gainlora,
  title={Gated integration of low-rank adaptation for continual learning of language models},
  author={Liang, Yan-Shuo and Li, Wu-Jun},
  year={2025}
}

@article{robustmerge,
  title={Robustmerge: Parameter-efficient model merging for mllms with direction robustness},
  author={Zeng, Fanhu and Guo, Haiyang and Zhu, Fei and Shen, Li and Tang, Hao},
  journal={arXiv preprint arXiv:2502.17159},
  year={2025}
}

@inproceedings{opcm,
  title={Merging on the Fly Without Retraining: A Sequential Approach to Scalable Continual Model Merging},
  author={Tang, Anke and Yang, Enneng and Shen, Li and Luo, Yong and Hu, Han and Zhang, Lefei and Du, Bo and Tao, Dacheng},
  booktitle={The Thirty-ninth Annual Conference on Neural Information Processing Systems},
  year={2025}
}

@inproceedings{acmap,
  title={Adapter merging with centroid prototype mapping for scalable class-incremental learning},
  author={Fukuda, Takuma and Kera, Hiroshi and Kawamoto, Kazuhiko},
  booktitle={Proceedings of the computer vision and pattern recognition conference},
  pages={4884--4893},
  year={2025}
}

@article{became,
  title={BECAME: Bayesian continual learning with adaptive model merging},
  author={Li, Mei and Lu, Yuxiang and Dai, Qinyan and Huang, Suizhi and Ding, Yue and Lu, Hongtao},
  journal={arXiv preprint arXiv:2504.02666},
  year={2025}
}

@inproceedings{seca,
  title={Harnessing textual semantic priors for knowledge transfer and refinement in clip-driven continual learning},
  author={He, Lingfeng and Cheng, De and Xu, Di and Wang, Huaijie and Wang, Nannan},
  booktitle={Proceedings of the AAAI Conference on Artificial Intelligence},
  volume={40},
  number={26},
  pages={21645--21653},
  year={2026}
}

@article{ekpc,
  title={EKPC: Elastic Knowledge Preservation and Compensation for Class-Incremental Learning: H. Wang et al.},
  author={Wang, Huaijie and Cheng, De and He, Lingfeng and Li, Yan and Li, Jie and Wang, Nannan and Gao, Xinbo},
  journal={International Journal of Computer Vision},
  volume={134},
  number={5},
  pages={238},
  year={2026},
  publisher={Springer}
}

@article{ewc-lora,
  title={Revisiting Weight Regularization for Low-Rank Continual Learning},
  author={Zheng, Yaoyue and Zhang, Yin and van de Weijer, Joost and van de Ven, Gido M and Du, Shaoyi and Zhang, Xuetao and Tian, Zhiqiang},
  journal={arXiv preprint arXiv:2602.17559},
  year={2026}
}

@article{saef,
  title={From Isolation to Integration: Building an Adaptive Expert Forest for Pre-Trained Model-based Class-Incremental Learning},
  author={Liu, Ruiqi and Diao, Boyu and Liu, Hangda and An, Zhulin and Wang, Fei and Xu, Yongjun},
  journal={arXiv preprint arXiv:2602.20911},
  year={2026}
}

@inproceedings{bilora,
  title={BiLoRA: almost-orthogonal parameter spaces for continual learning},
  author={Zhu, Hao and Zhang, Yifei and Dong, Junhao and Koniusz, Piotr},
  booktitle={Proceedings of the IEEE/CVF Conference on Computer Vision and Pattern Recognition},
  pages={25613--25622},
  year={2025}
}

@article{sam,
  title={Sharpness-aware minimization for efficiently improving generalization},
  author={Foret, Pierre and Kleiner, Ariel and Mobahi, Hossein and Neyshabur, Behnam},
  journal={arXiv preprint arXiv:2010.01412},
  year={2020}
}

@article{ewc,
  title={Overcoming catastrophic forgetting in neural networks},
  author={Kirkpatrick, James and Pascanu, Razvan and Rabinowitz, Neil and Veness, Joel and Desjardins, Guillaume and Rusu, Andrei A and Milan, Kieran and Quan, John and Ramalho, Tiago and Grabska-Barwinska, Agnieszka and others},
  journal={Proceedings of the national academy of sciences},
  volume={114},
  number={13},
  pages={3521--3526},
  year={2017},
  publisher={National Academy of Sciences}
}

@inproceedings{rapf,
  title={Class-incremental learning with clip: Adaptive representation adjustment and parameter fusion},
  author={Huang, Linlan and Cao, Xusheng and Lu, Haori and Liu, Xialei},
  booktitle={European Conference on Computer Vision},
  pages={214--231},
  year={2024},
  organization={Springer}
}

@article{merge1,
  title={Model merging with svd to tie the knots},
  author={Stoica, George and Ramesh, Pratik and Ecsedi, Boglarka and Choshen, Leshem and Hoffman, Judy},
  journal={arXiv preprint arXiv:2410.19735},
  year={2024}
}

@article{lora,
  title={Lora: Low-rank adaptation of large language models.},
  author={Hu, Edward J and Shen, Yelong and Wallis, Phillip and Allen-Zhu, Zeyuan and Li, Yuanzhi and Wang, Shean and Wang, Liang and Chen, Weizhu and others},
  journal={Iclr},
  volume={1},
  number={2},
  pages={3},
  year={2022}
}

@article{cl1,
  title={Class-incremental learning: survey and performance evaluation on image classification},
  author={Masana, Marc and Liu, Xialei and Twardowski, Bart{\l}omiej and Menta, Mikel and Bagdanov, Andrew D and Van De Weijer, Joost},
  journal={IEEE Transactions on Pattern Analysis and Machine Intelligence},
  volume={45},
  number={5},
  pages={5513--5533},
  year={2022},
  publisher={IEEE}
}

@incollection{cl3,
  title={Catastrophic interference in connectionist networks: The sequential learning problem},
  author={McCloskey, Michael and Cohen, Neal J},
  booktitle={Psychology of learning and motivation},
  volume={24},
  pages={109--165},
  year={1989},
  publisher={Elsevier}
}

@article{ptm,
  title={An image is worth 16x16 words: Transformers for image recognition at scale},
  author={Dosovitskiy, Alexey and Beyer, Lucas and Kolesnikov, Alexander and Weissenborn, Dirk and Zhai, Xiaohua and Unterthiner, Thomas and Dehghani, Mostafa and Minderer, Matthias and Heigold, Georg and Gelly, Sylvain and others},
  journal={arXiv preprint arXiv:2010.11929},
  year={2020}
}

@inproceedings{-r,
  title={The many faces of robustness: A critical analysis of out-of-distribution generalization},
  author={Hendrycks, Dan and Basart, Steven and Mu, Norman and Kadavath, Saurav and Wang, Frank and Dorundo, Evan and Desai, Rahul and Zhu, Tyler and Parajuli, Samyak and Guo, Mike and others},
  booktitle={Proceedings of the IEEE/CVF international conference on computer vision},
  pages={8340--8349},
  year={2021}
}

@inproceedings{-a,
  title={Natural adversarial examples},
  author={Hendrycks, Dan and Zhao, Kevin and Basart, Steven and Steinhardt, Jacob and Song, Dawn},
  booktitle={Proceedings of the IEEE/CVF conference on computer vision and pattern recognition},
  pages={15262--15271},
  year={2021}
}

@article{cifar100,
  title={Learning multiple layers of features from tiny images},
  author={Krizhevsky, Alex and Hinton, Geoffrey and others},
  year={2009},
  publisher={Toronto, ON, Canada}
}

@article{cub,
  title={The caltech-ucsd birds-200-2011 dataset},
  author={Wah, Catherine and Branson, Steve and Welinder, Peter and Perona, Pietro and Belongie, Serge},
  year={2011}
}

@inproceedings{domainnet,
  title={Moment matching for multi-source domain adaptation},
  author={Peng, Xingchao and Bai, Qinxun and Xia, Xide and Huang, Zijun and Saenko, Kate and Wang, Bo},
  booktitle={Proceedings of the IEEE/CVF international conference on computer vision},
  pages={1406--1415},
  year={2019}
}

@article{imagenet-21k,
  title={Imagenet large scale visual recognition challenge},
  author={Russakovsky, Olga and Deng, Jia and Su, Hao and Krause, Jonathan and Satheesh, Sanjeev and Ma, Sean and Huang, Zhiheng and Karpathy, Andrej and Khosla, Aditya and Bernstein, Michael and others},
  journal={International journal of computer vision},
  volume={115},
  number={3},
  pages={211--252},
  year={2015},
  publisher={Springer}
}

@article{CoOp,
  title={Learning to prompt for vision-language models},
  author={Zhou, Kaiyang and Yang, Jingkang and Loy, Chen Change and Liu, Ziwei},
  journal={International Journal of Computer Vision},
  volume={130},
  number={9},
  pages={2337--2348},
  year={2022},
  publisher={Springer}
}

@inproceedings{zsclip,
  title={Learning transferable visual models from natural language supervision},
  author={Radford, Alec and Kim, Jong Wook and Hallacy, Chris and Ramesh, Aditya and Goh, Gabriel and Agarwal, Sandhini and Sastry, Girish and Askell, Amanda and Mishkin, Pamela and Clark, Jack and others},
  booktitle={International conference on machine learning},
  pages={8748--8763},
  year={2021},
  organization={PmLR}
}

@article{simplecl,
  title={Revisiting class-incremental learning with pre-trained models: Generalizability and adaptivity are all you need},
  author={Zhou, Da-Wei and Cai, Zi-Wen and Ye, Han-Jia and Zhan, De-Chuan and Liu, Ziwei},
  journal={International Journal of Computer Vision},
  pages={1--21},
  year={2024},
  publisher={Springer}
}

@article{PROOF-TPAMI25,
  title={Learning without forgetting for vision-language models},
  author={Zhou, Da-Wei and Zhang, Yuanhan and Wang, Yan and Ning, Jingyi and Ye, Han-Jia and Zhan, De-Chuan and Liu, Ziwei},
  journal={IEEE Transactions on Pattern Analysis and Machine Intelligence},
  year={2025},
  publisher={IEEE}
}

@inproceedings{cars,
  title={3d object representations for fine-grained categorization},
  author={Krause, Jonathan and Stark, Michael and Deng, Jia and Fei-Fei, Li},
  booktitle={Proceedings of the IEEE international conference on computer vision workshops},
  pages={554--561},
  year={2013}
}

@article{ucf101_dataset,
  title={Ucf101: A dataset of 101 human actions classes from videos in the wild},
  author={Soomro, Khurram and Zamir, Amir Roshan and Shah, Mubarak},
  journal={arXiv preprint arXiv:1212.0402},
  year={2012}
}

@article{aircraft_dataset,
  title={Fine-grained visual classification of aircraft},
  author={Maji, Subhransu and Rahtu, Esa and Kannala, Juho and Blaschko, Matthew and Vedaldi, Andrea},
  journal={arXiv preprint arXiv:1306.5151},
  year={2013}
}

@inproceedings{vptnspNIPS-24,
  title = {Visual {{Prompt Tuning}} in {{Null Space}} for {{Continual Learning}}},
  booktitle = {Conference on {{Neural Information Processing Systems}}},
  author = {Lu, Yue and Zhang, Shizhou and Cheng, De and Xing, Yinghui and Wang, Nannan and Wang, Peng and Zhang, Yanning},
  year = 2024
}

@inproceedings{vptcpg-AAAI25,
  title={Training Consistent Mixture-of-Experts-Based Prompt Generator for Continual Learning},
  author={Lu, Yue and Zhang, Shizhou and Cheng, De and Liang, Guoqiang and Xing, Yinghui and Wang, Nannan and Zhang, Yanning},
  booktitle={Proceedings of the AAAI Conference on Artificial Intelligence},
  volume={39},
  number={18},
  pages={19152--19160},
  year={2025}
}

@inproceedings{
loda2026,
title={Task-Driven Subspace Decomposition for Knowledge Sharing and Isolation in LoRA-based Continual Learning},
author={Lingfeng He and De Cheng and Huaijie Wang and Xiaofeng Zhu and Xi Yang and Nannan Wang and Xinbo Gao},
booktitle={Forty-third International Conference on Machine Learning},
year={2026}
}

@inproceedings{DPR,
  title={Disentangled prompt representation for domain generalization},
  author={Cheng, De and Xu, Zhipeng and Jiang, Xinyang and Wang, Nannan and Li, Dongsheng and Gao, Xinbo},
  booktitle={Proceedings of the IEEE/CVF Conference on Computer Vision and Pattern Recognition},
  pages={23595--23604},
  year={2024}
}

@inproceedings{PAPT,
  title={Adversarial domain prompt tuning and generation for single domain generalization},
  author={Xu, Zhipeng and Cheng, De and Jiang, Xinyang and Wang, Nannan and Li, Dongsheng and Gao, Xinbo},
  booktitle={Proceedings of the IEEE/CVF Conference on Computer Vision and Pattern Recognition},
  pages={18584--18595},
  year={2025}
}

@article{PADG,
  title={Prompt disentanglement via language guidance and representation alignment for domain generalization},
  author={Cheng, De and Xu, Zhipeng and Jiang, Xinyang and Li, Dongsheng and Wang, Nannan and Gao, Xinbo},
  journal={IEEE Transactions on Pattern Analysis and Machine Intelligence},
  year={2026},
  publisher={IEEE}
}

@article{RD-MLDG,
  title={Reasoning-Driven Multimodal LLM for Domain Generalization},
  author={Xu, Zhipeng and Wang, Zilong and Jiang, Xinyang and Li, Dongsheng and Cheng, De and Wang, Nannan},
  journal={arXiv preprint arXiv:2602.23777},
  year={2026}
}

@inproceedings{chenginterference,
  title={Interference-Isolated Elastic Weight Consolidation and Knowledge Calibration for Incremental Object Detection},
  author={Cheng, De and Zeng, Mingyue and Xu, Zhipeng and Xu, Di and Wang, Nannan and Gao, Xinbo},
  booktitle={The Fourteenth International Conference on Learning Representations}
}

@inproceedings{wang2026stpr,
  title     = {StPR: Spatiotemporal Preservation and Routing for Exemplar-Free Video Class-Incremental Learning},
  author    = {Wang, Huaijie and Cheng, De and Li, Guozhang and Xu, Zhipeng and He, Lingfeng and Li, Jie and Wang, Nannan and Gao, Xinbo},
  booktitle = {The Fourteenth International Conference on Learning Representations},
  year      = {2026},
  url       = {https://openreview.net/forum?id=VAn2YVMuZC}
}

@article{PAPT++,
  title={PAPT++: Risk-Aware Adversarial Tuning and Generation for Single Domain Generalization},
  author={Xu, Zhipeng and Cheng, De and Jiang, Xinyang and He, Lingfeng and Wang, Huaijie and Li, Dongsheng and Wang, Nannan and Gao, Xinbo},
  journal={arXiv preprint arXiv:2609.04837},
  year={2026}
}

@article{DCD,
  title={Dual-Branch Cross-Projection Debiasing through Diffusion-based Disentanglement},
  author={Zhao, Xiangqian and Jiang, Xinyang and Xu, Zhipeng and He, Lingfeng and Wang, Zilong and Li, Dongsheng and Cheng, De and Wang, Nannan},
  journal={arXiv preprint arXiv:2606.24161},
  year={2026}
}

@inproceedings{SIKD,
  title     = {Symbiosis-Inspired Knowledge Distillation for Incremental Object Detection},
  author    = {Zeng, Mingyue and Cheng, De and Xu, Zhipeng and Wang, Huaijie and Wang, Nannan and Gao, Xinbo},
  booktitle = {Proceedings of the 43rd International Conference on Machine Learning},
  year      = {2026}
}

@inproceedings{li2026few,
  title={Few-Shot Hybrid Incremental Learning: Continually Learning under Data Scarcity and Task Uncertainty},
  author={Li, Yan and Shi, Yuzhu and Zhou, Kan and Zhang, Shu and He, Diqi and Zhang, Dingwen and Han, Junwei},
  booktitle={Proceedings of the IEEE/CVF Conference on Computer Vision and Pattern Recognition},
  pages={32334--32344},
  year={2026}
}

\clearpage
\appendix

\counterwithin{equation}{subsection}
\renewcommand{\theequation}{\thesubsection.\arabic{equation}}

\section{Training Pipeline}
\label{training}

\begin{figure*}[htbp]
\centering
\includegraphics[width=1\textwidth]{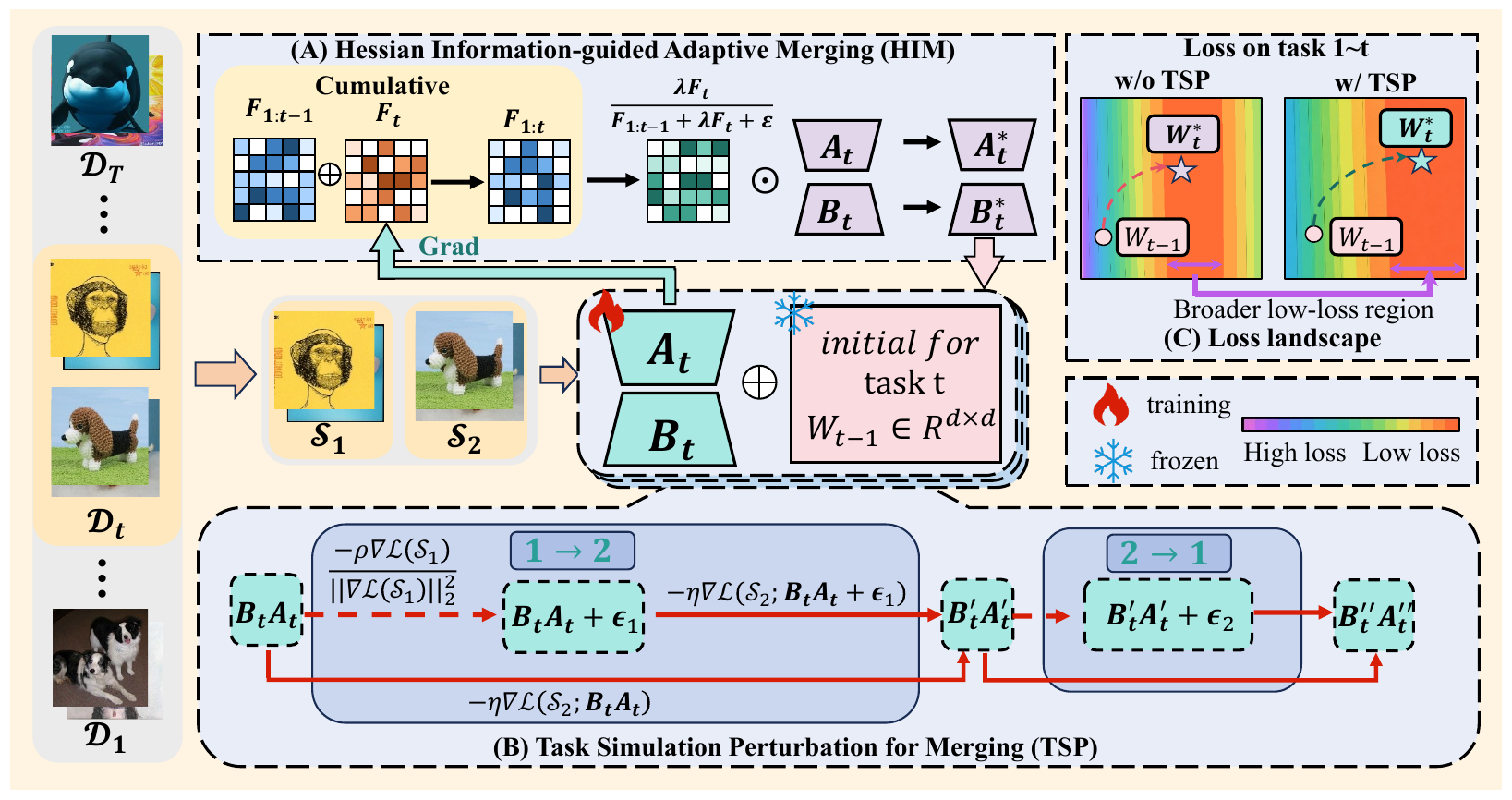}
\caption{
Overall framework of $\mathrm{M}^{2}\mathrm{LE}$.
(B) During training, TSP simulates cross task interference
between label-disjoint subsets.
(A) After training, HIM adaptively merges the LoRA update
using historical and current-task Fisher information.
}
\label{fig:framework}
\end{figure*}

Figure~\ref{fig:framework} illustrates the overall framework of
$\mathrm{M}^{2}\mathrm{LE}$, and the training pipeline of $\text{M}^2$LE is outlined in Algorithm~\ref{alg:liam}.
At task $t$, we start from the inference parameter $\mathbf{W}_{t-1}$ and adapt the frozen backbone on the current data $\mathcal{D}_t$ with LoRA, where the current update is $\Delta\mathbf{W}_t=\mathbf{B}_t\mathbf{A}_t$.
During current task training, TSP simulates cross-task interference with label-disjoint subsets, encouraging the learned update to remain stable for subsequent merging.
After convergence, HIM estimates the current task Fisher $\mathbf{F}_t$ and obtains the target merged update $\Delta\mathbf{W}_t^*$ by Eq.~\eqref{eq:element_wise_solution}.
The merged update is then integrated into the inference model, and the Fisher statistics are accumulated for later tasks.

\begin{algorithm}[h]
\caption{Training pipeline of the proposed \underline{M}erging Strength \underline{M}odulation via Joint \underline{L}oss \underline{E}stimation ($\text{M}^2$LE).}
\label{alg:liam}
\begin{algorithmic}[1]
\Statex \textbf{Input:} Task sequence $\{\mathcal{D}_{1}, \ldots, \mathcal{D}_{T}\}$, where 
$\mathcal{D}_{t}=\{(x_t^i,y_t^i)\}_{i=1}^{n_t}$; ViT model $f(\cdot)$ with weight $\mathbf{W}_{0}$; 
LoRA $\Delta\mathbf{W}_t=\mathbf{B}_t\mathbf{A}_t$ in each layer; accumulated Fisher statistics $\mathbf{F}_{1:t-1}$ in each layer.
\Statex \textbf{Output:} The adapted inference parameters $\mathbf{W}_{t}$ for each layer in the ViT model.

\State Initialization: $\mathbf{F}_{1:t-1} \gets \mathbf{0}^{d_o\times d_i}$.
\For{task $t=1$ to $T$}
    \State Freeze the inference weight $\mathbf{W}_{t-1}$ and attach current-task LoRA modules.
    \For{each LoRA layer}
        \State Initialize current-task LoRA parameters $\mathbf{A}_t$ and $\mathbf{B}_t$.
    \EndFor

    \BeginBox[phaseduring]
    \LComment{During Training: Task Simulation Perturbation for Merging}
    \Repeat
        \State Sample a mini-batch $\mathcal{S}$ from $\mathcal{D}_{t}$ and split it into two label-disjoint subsets $\mathcal{S}_1$ and $\mathcal{S}_2$;
        \State Update trainable parameters with the dual TSP objective in Eq.~\eqref{eq:meta_update_ab} and Eq.~\eqref{eq:meta_update_ba}.
    \Until{Convergence on $\mathcal{D}_{t}$}
    \EndBox
    \BeginBox[phaseafter]
    \LComment{After Training: Hessian Information-guided Adaptive Merging}
    \For{each LoRA layer}
        \State Given the trained update $\Delta\mathbf{W}_t$, estimate $\mathbf{F}_t$ on $\mathcal{D}_t$ and compute the HIM merged update $\Delta\mathbf{W}_t^{*}$ by Eq.~\eqref{eq:element_wise_solution}.
        \State Update the inference model weight $
        \mathbf{W}_{t} \gets \mathbf{W}_{t-1}+\Delta\mathbf{W}_{t}^{*}$;
        \State Update accumulated Fisher $
        \mathbf{F}_{1:t}\gets \mathbf{F}_{1:t-1}+\mathbf{F}_{t}.
        $;
    \EndBox
    \EndFor

\EndFor
\end{algorithmic}
\end{algorithm}

\section{Theoretical Analysis}
\label{merge}

Starting from Eq.~\eqref{eq:quad_obj_main}, we solve the following quadratic problem over the candidate merged update $\Delta \hat{\mathbf{w}}_t$:
\begin{equation}
\begin{aligned}
\Delta \mathbf{w}_t^*
=
\arg\min_{\Delta \hat{\mathbf{w}}_t} \quad
& \sum_{i=1}^{t-1} 
\frac{1}{2}
\Delta \hat{\mathbf{w}}_t^{\top}
\mathbf{H}_i
\Delta \hat{\mathbf{w}}_t  \\
&+
\frac{\lambda}{2}
(\Delta \hat{\mathbf{w}}_t-\Delta \mathbf{w}_t)^{\top}
\mathbf{H}_t
(\Delta \hat{\mathbf{w}}_t-\Delta \mathbf{w}_t).
\end{aligned}
\label{eq:app_jem_quad}
\end{equation}

We first expand the current task term and separate the parts that depend on $\Delta \hat{\mathbf{w}}_t$:
\begin{equation}
\begin{aligned}
&\frac{\lambda}{2}
(\Delta \hat{\mathbf{w}}_t-\Delta \mathbf{w}_t)^{\top}
\mathbf{H}_t
(\Delta \hat{\mathbf{w}}_t-\Delta \mathbf{w}_t)
\\
=&
\frac{\lambda}{2}
\Delta \hat{\mathbf{w}}_t^{\top}
\mathbf{H}_t
\Delta \hat{\mathbf{w}}_t
-
\lambda
\Delta \hat{\mathbf{w}}_t^{\top}
\mathbf{H}_t
\Delta \mathbf{w}_t
+
\frac{\lambda}{2}
\Delta \mathbf{w}_t^{\top}
\mathbf{H}_t
\Delta \mathbf{w}_t .
\end{aligned}
\label{eq:app_expand_current}
\end{equation}
The last term in Eq.~\eqref{eq:app_expand_current} is constant with respect to $\Delta \hat{\mathbf{w}}_t$ and can be ignored when solving the minimization problem. 
Thus, the derivative of Eq.~\eqref{eq:app_jem_quad} with respect to $\Delta \hat{\mathbf{w}}_t$ is:
\begin{equation}
\begin{aligned}
\frac{\partial}{\partial \Delta \hat{\mathbf{w}}_t}
&\left[
\sum_{i=1}^{t-1}
\frac{1}{2}
\Delta \hat{\mathbf{w}}_t^{\top}
\mathbf{H}_i
\Delta \hat{\mathbf{w}}_t
+
\frac{\lambda}{2}
\Delta \hat{\mathbf{w}}_t^{\top}
\mathbf{H}_t
\Delta \hat{\mathbf{w}}_t
-
\lambda
\Delta \hat{\mathbf{w}}_t^{\top}
\mathbf{H}_t
\Delta \mathbf{w}_t
\right]
\\
&=
\sum_{i=1}^{t-1}
\mathbf{H}_i
\Delta \hat{\mathbf{w}}_t
+
\lambda
\mathbf{H}_t
\Delta \hat{\mathbf{w}}_t
-
\lambda
\mathbf{H}_t
\Delta \mathbf{w}_t .
\end{aligned}
\label{eq:app_jem_gradient}
\end{equation}
By setting Eq.~\eqref{eq:app_jem_gradient} to zero and denoting the solution by $\Delta \mathbf{w}_t^*$, we obtain:
\begin{equation}
\sum_{i=1}^{t-1}
\mathbf{H}_i
\Delta \mathbf{w}_t^*
+
\lambda
\mathbf{H}_t
\Delta \mathbf{w}_t^*
-
\lambda
\mathbf{H}_t
\Delta \mathbf{w}_t
=
\mathbf{0}.
\label{eq:app_stationary}
\end{equation}
Rearranging the terms gives the linear system, which corresponds to Eq.~\eqref{eq:closed_form_main}:
\begin{equation}
\left(
\sum_{i=1}^{t-1}
\mathbf{H}_i
+
\lambda \mathbf{H}_t
\right)
\Delta \mathbf{w}_t^*
=
\lambda
\mathbf{H}_t
\Delta \mathbf{w}_t.
\label{eq:app_linear_system}
\end{equation}



\section{More Detailed Experimental Results}

\subsection{Forgetting metric}
\label{forget}
We measure forgetting after task $t \geq 1$ as
$\mathrm{Forgetting}_t = \frac{1}{t}\sum_{i=0}^{t-1}
\left(\max_{j\in\{i,\ldots,t-1\}} a_{j,i} - a_{t,i}\right)$,
where $a_{j,i}$ denotes the accuracy on task $i$ after learning
task $j$.

\subsection{Effect of Different LoRA Ranks on $\text{M}^2$LE Performance}
\label{rank}
\begin{table*}[h]
\centering
\caption{\footnotesize{Analysis of the effect of rank on additional parameters, storage and computational overhead, and performance on 20S-ImageNet-R. We compare our $\text{M}^2$LE with different ranks against representative CL methods.}}\label{tab: memory_overhead_appx}
\begin{adjustbox}{max width=1.00\textwidth}
\begin{tabular}{
lccccccccc
}\toprule
\multirow{2}*{Methods} & \multirow{2}*{Venue} & \multicolumn{3}{c}{Training Stage (Per Sample)} & 
\multicolumn{2}{c}{Inference Stage (Per Sample)}
& \multicolumn{2}{c}{20S-ImageNet-R} \\
\cmidrule{3-9}
& & \makecell{Extra\\Memory} & \makecell{Trainable\\Params} & \makecell{Training \\ GFLOPs} & \makecell{Extra\\Params} & \makecell{Inference \\ GFLOPs} & 
$\mathcal{A}_{Last}$ $\uparrow$ & $\mathcal{A}_{Avg}$ \\
\midrule
\multicolumn{5}{l}{\textbf{Methods based on the ViT-21K model (86M).}} \\
\midrule
CODA \cite{codaprompt} & CVPR-23 &  15.4MB & 3.84M (4.46\%) & 35.17 & 3.84M (4.46\%) & 35.17 & $69.96_{\pm 0.50}$ & $75.34_{\pm 0.85}$ \\
SLCA~\cite{slca} & ICCV-23 & 450.6MB & 86M (100\%) & 17.60 & 0.00M (0.00\%) & 17.60 & $74.63_{\pm 1.55}$& $79.92_{\pm 1.29} $ \\
SSIAT~\cite{ssiat} & CVPR-24 & 450.6MB & 1.19M (1.38\%) & 17.81 & 1.19M (1.38\%) & 17.81 & $75.67_{\pm 0.24}$ & ${82.30}_{\pm 0.36}$ \\
InfLoRA~\cite{inflora} & CVPR-24 & 11.3MB & 0.37M (0.43\%) & 17.67 & 0.00M (0.00\%) & 17.67 & $71.01_{\pm 0.45}$ & $77.28_{\pm 0.45}$\\
VPT-NSP~\cite{vptnspNIPS-24} & NIPS-24 & 28.4MB & 0.09M (0.10\%) & 17.72 & 0.09M (0.10\%) & 17.72 & $75.69_{\pm 0.61}$ & $81.87_{\pm 0.59}$ \\
SD-LoRA~\cite{sdlora} & ICLR-25 & 14.1MB & 0.37M (0.43\%) & 20.51 & 0.00M (0.00\%) & 17.72 & $75.26_{\pm 0.37}$ & $80.22_{\pm 0.72}$\\
VPT-CPG~\cite{vptcpg-AAAI25} & AAAI-25 & 28.4MB & 1.66M (1.93\%) & 17.72 & 1.66M (1.93\%) & 17.72 & $75.33_{\pm 0.37}$ & $80.22_{\pm 0.72}$ \\
\midrule
\textbf{$\textbf{M}^2$LE--r=4 (Ours)} & -- & 54.0MB & 0.15M (0.17\%) & 17.59 & 0.00M (0.00\%) & 17.60 & $\mathbf{76.92_{\pm 0.20}}$ & $\mathbf{83.28_{\pm 0.25}}$\\
\textbf{$\textbf{M}^2$LE--r=8 (Ours)} & -- & 54.0MB & 0.29M (0.34\%) & 17.62 & 0.00M (0.00\%) & 17.60 & $\mathbf{77.32_{\pm 0.55}}$ & $\mathbf{83.68_{\pm 0.51}}$\\
\textbf{$\textbf{M}^2$LE--r=16 (Ours)} & -- & 54.0MB & 0.59M (0.69\%) & 17.68 & 0.00M (0.00\%) & 17.60 & $\mathbf{78.01_{\pm 0.13}}$ & $\mathbf{84.07_{\pm 0.43}}$ \\
\textbf{$\textbf{M}^2$LE--r=32 (Ours)} & -- & 54.0MB & 1.18M (1.37\%) & 17.80 & 0.00M (0.00\%) & 17.60 & $\mathbf{78.62_{\pm 0.43}}$ & $\mathbf{84.62_{\pm 0.26}}$ \\
\textbf{$\textbf{M}^2$LE--r=64 (Ours)} & -- & 54.0MB & 2.36M (2.74\%) & 18.03 & 0.00M (0.00\%) & 17.60 & $\mathbf{78.20_{\pm 0.30}}$ & $\mathbf{84.66_{\pm 0.70}}$ \\
\bottomrule
\end{tabular}
\end{adjustbox}
\end{table*}

We study the effect of rank in $\text{M}^2$LE on 20S-ImageNet-R in Tab.~\ref{tab: memory_overhead_appx}. 
We vary the LoRA rank $r \in \{4, 8, 16, 32, 64\}$ and denote each variant as $\text{M}^2$LE-r=$r$. 
Across all ranks, $\text{M}^2$LE achieves competitive performance with only a small number of trainable parameters, ranging from $0.15$M to $2.36$M. 
It also keeps the inference overhead unchanged, with $0.00$M extra parameters and $17.60$ GFLOPs, since the learned components are merged into the backbone after training. 
For each LoRA layer, the maintained Fisher statistic $\mathbf{F}_{1:t}$ has size $d_o \times d_i$, so the extra storage remains $54.0$MB for all ranks once the layer dimensions are fixed.

As $r$ increases from $4$ to $32$, $\mathcal{A}_{Last}$ improves from $76.92$ to $78.62$, while $\mathcal{A}_{Avg}$ increases from $83.28$ to $84.62$. 
Further increasing the rank to $64$ brings only marginal improvement in $\mathcal{A}_{Avg}$ ($84.66$) but slightly decreases $\mathcal{A}_{Last}$ to $78.20$, together with higher training cost. 
Thus, we adopt $r=32$ as a practical trade-off between accuracy and efficiency in the remaining experiments.

\begin{table}[t]
\centering
\caption{Comparison of different HIM variants on 10S-ImageNet-R and 10S-ImageNet-A.}
\label{tab:old_task}
\begin{adjustbox}{max width=0.85\linewidth}
\begin{tabular}{lcc|cc}
\toprule
\multirow{2}{*}{Variant} 
& \multicolumn{2}{c|}{10S-ImageNet-R} 
& \multicolumn{2}{c}{10S-ImageNet-A} \\
& $\mathcal{A}_{Last}\uparrow$ & $\mathcal{A}_{Avg}\uparrow$
& $\mathcal{A}_{Last}\uparrow$ & $\mathcal{A}_{Avg}\uparrow$ \\
\midrule
HIM-last ($k=1$)
& $76.23_{\pm0.76}$ & $83.81_{\pm0.61}$ 
& $58.17_{\pm2.38}$ & $68.84_{\pm1.93}$ \\
HIM-$k=3$
& $78.88_{\pm0.62}$ & $85.37_{\pm0.50}$ 
& $60.81_{\pm1.10}$ & $69.97_{\pm2.28}$ \\
HIM-$k=5$
& $80.04_{\pm0.35}$ & $85.73_{\pm0.47}$ 
& $60.92_{\pm0.97}$ & $70.14_{\pm2.48}$ \\
HIM-$k=7$
& $80.21_{\pm0.45}$ & $85.77_{\pm0.45}$ 
& $61.16_{\pm1.15}$ & $70.17_{\pm2.42}$ \\
\textbf{HIM-full (Ours)}
& $\mathbf{80.36}_{\bm{\pm}\mathbf{0.40}}$ & $\mathbf{85.78}_{\bm{\pm}\mathbf{0.44}}$ 
& $\mathbf{61.35}_{\bm{\pm}\mathbf{1.06}}$ & $\mathbf{70.23}_{\bm{\pm}\mathbf{2.04}}$ \\
\bottomrule
\end{tabular}
\end{adjustbox}
\end{table}

\subsection{Analysis of HIM Variants}
\label{more_old}

We further analyze several HIM variants that retain different numbers of past tasks in the old task term.
For a more explicit comparison, we define a truncated HIM variant by retaining only the most recent $k$ past tasks in the old task loss impact:
\begin{equation}
\mathcal{J}_{\mathrm{old}}^{(k)}(\hat{\Delta\mathbf{W}}_t)
=
\sum_{i=\max(1,t-k)}^{t-1}
\left(
\mathcal{L}_i(\mathbf{W}_{t-1}+\hat{\Delta\mathbf{W}}_t)
-
\mathcal{L}_i(\mathbf{W}_{t-1})
\right).
\label{eq:truncated_old_task_term}
\end{equation}
Here, $k \in \{1,3,5,7\}$ controls the number of past tasks used in the old task term. 
Correspondingly, this variant replaces the accumulated Fisher statistic in Eq.~\eqref{eq:element_wise_solution} with $\mathbf{F}_{t-k:t-1}=\sum_{i=\max(1,t-k)}^{t-1}\mathbf{F}_i$, while full HIM uses $\mathbf{F}_{1:t-1}=\sum_{i=1}^{t-1}\mathbf{F}_i$. 
When $k=1$, the variant uses only task $t-1$, corresponding to the HIM-last setting discussed in the main text.


As shown in Tab.~\ref{tab:old_task}, HIM-last gives much lower performance, i.e., $76.23/83.81$ on 10S-ImageNet-R and $58.17/68.84$ on 10S-ImageNet-A. 
When more past tasks are retained, the performance consistently improves, e.g., HIM-$k=5$ reaches $80.04/85.73$ and $60.92/70.14$ on the two datasets, respectively. 
Full HIM achieves the best results of $80.36/85.78$ and $61.35/70.23$, indicating that using a broader set of past tasks provides more complete constraints for preserving previously learned knowledge during merging. 
These results also validate the use of $\mathbf{F}_{1:t-1}$ in HIM, which aggregates information from all past tasks rather than relying only on recent ones.

\subsection{Analysis of Historical Fisher Reuse}
\label{fisher}

We further analyze the accumulated Fisher statistics in HIM on 10S-ImageNet-R. 
HIM estimates each task's Fisher at its pre-merge solution while the data are available, and retains only the accumulated statistic $F_{1:t-1}$. 
For comparison, Recomputed Fisher revisits previous task data before each merge and re-estimates their accumulated Fisher at its current merged model.
The recomputed statistic is used for merging and subsequent learning, so the two methods follow their respective model trajectories. 
We measure their Fisher similarity by
$S_t = \frac{1}{|\mathcal{L}|}\sum_{\ell\in\mathcal{L}}
\cos(F_{1:t-1}^{\ell},\widetilde{F}_{1:t-1}^{\ell})$,
where $\mathcal{L}$ denotes the set of layers with Fisher statistics, and $F_{1:t-1}^{\ell}$ and
$\widetilde{F}_{1:t-1}^{\ell}$ are the accumulated statistics from HIM and Recomputed Fisher, respectively.

As shown in Tab.~\ref{tab:historical_fisher}, $S_t$ remains at least 0.986 across the evaluated stages, indicating similar directions of the accumulated Fisher statistics.
Recomputed Fisher achieves similar accuracy to HIM, with a maximum difference of 0.28 performance and the same final $\mathcal{A}_{Last}$ of 80.67\%. 
These results support using the accumulated historical Fisher in HIM, avoiding repeated access
to previous task data with little impact on performance over the evaluated task sequence.
\begin{table}[t]
    \centering
    \caption{Task-by-task comparison of accumulated historical
Fisher and recomputed Fisher on 10S-ImageNet-R. Results are obtained with a single random seed.}
    \label{tab:historical_fisher}
    \resizebox{\linewidth}{!}{
    \begin{tabular}{lcccccccccc}
        \toprule
        Metric
        & Task 0 & Task 1 & Task 2 & Task 3 & Task 4
        & Task 5 & Task 6 & Task 7 & Task 8 & Task 9 \\
        \midrule
        $S_t \uparrow$
        & -- & 1.000 & 0.988 & 0.986 & 0.986
        & 0.987 & 0.988 & 0.987 & 0.987 & 0.986 \\
        HIM $\mathcal{A}_{Last} \uparrow$
        & 96.66 & 92.81 & 89.41 & 87.39 & 84.69
        & 83.84 & 83.18 & 82.62 & 81.68 & 80.67 \\
        Recomputed Fisher $\mathcal{A}_{Last} \uparrow$
        & 96.66 & 92.81 & 89.36 & 87.11 & 84.76
        & 83.70 & 83.15 & 82.70 & 81.68 & 80.67 \\
        \bottomrule
    \end{tabular}
    }
\end{table}

\subsection{Hyper-Parameter Analysis}
\label{hyper}

\begin{figure*}[htbp]
\centering
\includegraphics[width=1\textwidth]{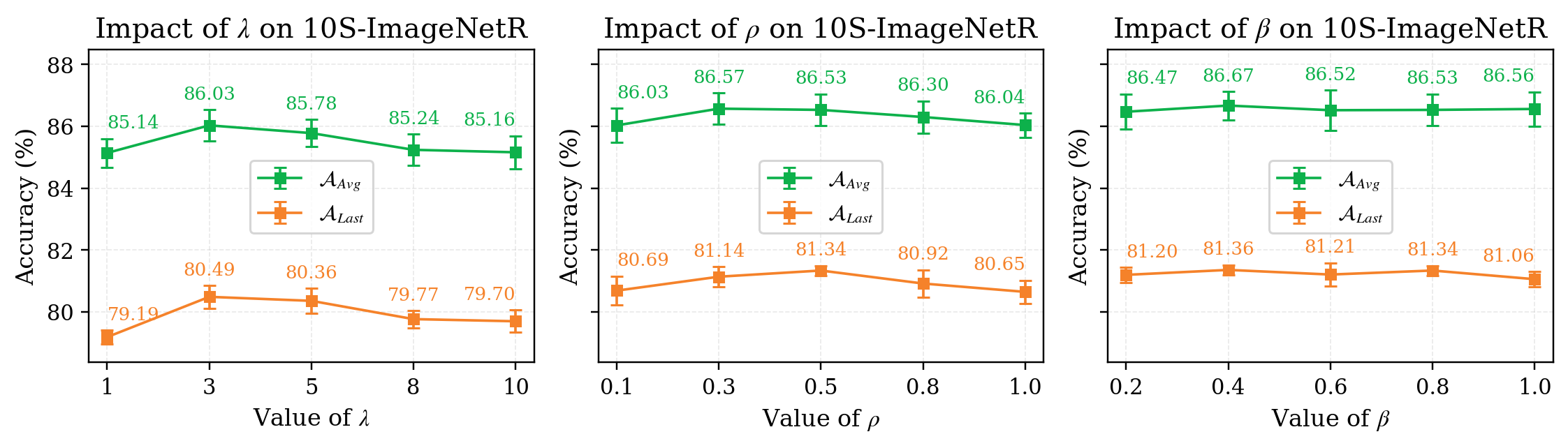}
\caption{Hyper-parameter analysis on 10S-ImageNet-R. From left to right, $\lambda$, $\rho$, and $\beta$ are varied while the other two are fixed. All results are averaged over three random seeds, and error bars denote the standard deviation.
}
\label{fig:hparam_r}
\end{figure*}

\begin{figure*}[htbp]
\centering
\includegraphics[width=1\textwidth]{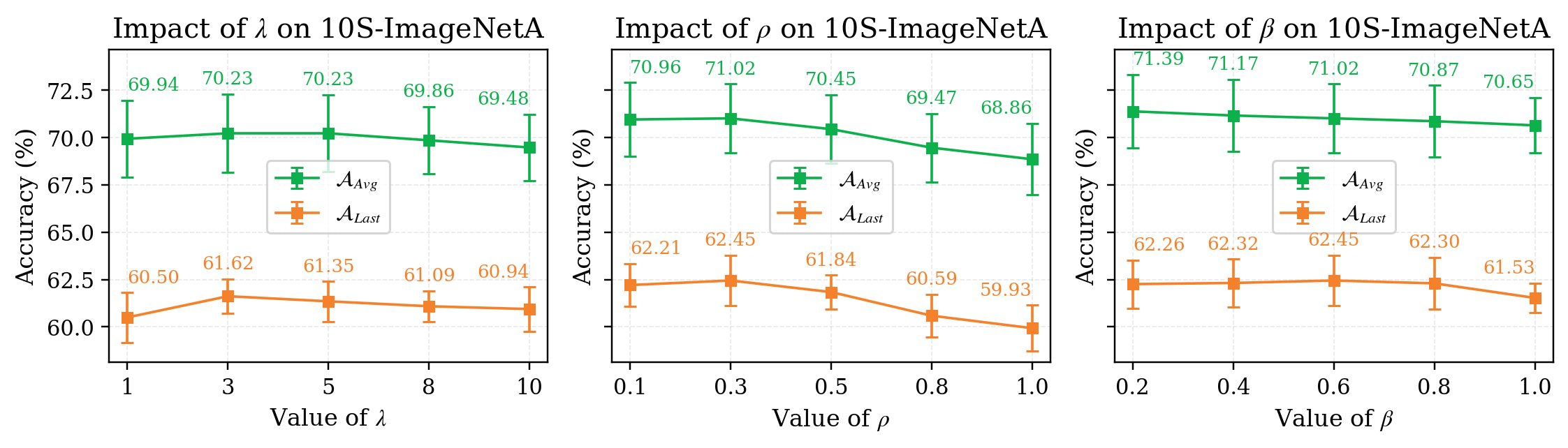}
\caption{Same as Fig.~\ref{fig:hparam_r}, but on 10S-ImageNet-A.
}
\label{fig:hparam_a}
\end{figure*}


We conduct experiments to investigate how the key hyper-parameters $\lambda$, $\rho$, and $\beta$ affect performance on 10S-ImageNet-R and 10S-ImageNet-A. 
Here, $\lambda$ is the trade-off coefficient in the merge objective in Eq.~\eqref{eq:element_wise_solution}, $\rho$ controls the perturbation magnitude in Eq.~\eqref{eq:meta_probe}, and $\beta$ balances the perturbed and clean losses in Eq.~\eqref{eq:meta_mix}. 
We use the default settings $\lambda=5$, $\rho=0.5$, and $\beta=0.8$ on ImageNet-R, and $\lambda=5$, $\rho=0.3$, and $\beta=0.6$ on ImageNet-A.
For each analysis, we vary only one hyper-parameter while fixing the other two at their default values. 
The results are shown in Fig.~\ref{fig:hparam_r} and Fig.~\ref{fig:hparam_a}. 
All results are averaged over three random seeds, and we report the mean with error bars.

\textbf{Impact of $\lambda$.}
The hyper-parameter $\lambda$ controls the relative weight of the current task term in the merge objective, and thus adjusts the balance between old task retention and current task adaptation.
A too-small $\lambda$ (e.g., $\lambda=1$) underweights the current task term and limits new task adaptation, whereas an overly large $\lambda$ (e.g., $\lambda \geq 10$) makes the merged update overly biased toward the current task and weakens old task retention.
Overall, the results remain stable for $1 \leq \lambda \leq 10$, with the best trade-off achieved around $\lambda=3$--$5$.
Based on this observation, we set $\lambda=5$.

\textbf{Impact of $\rho$.}
The hyper-parameter $\rho$ controls the perturbation magnitude in the TSP.
A too-small $\rho$ (e.g., $\rho=0.1$) leads to a weak cross-subset probing effect, whereas a too-large $\rho$ (e.g., $\rho \geq 0.8$) over-amplifies the perturbation.
Overall, the results remain stable for $0.1 \leq \rho \leq 1.0$, with the best performance achieved at moderate values.
Accordingly, we set $\rho=0.5$ on ImageNet-R and $\rho=0.3$ on ImageNet-A.

\textbf{Impact of $\beta$.}
The hyper-parameter $\beta$ controls the relative weight of the perturbed loss and the clean loss in the TSP.
A too-small $\beta$ (e.g., $\beta=0.2$) underweights the perturbation signal, whereas a too-large $\beta$ (e.g., $\beta=1.0$) makes the perturbation effect overly strong and is less favorable overall.
Overall, the results remain stable for $0.2 \leq \beta \leq 1.0$, and the best results are obtained at moderate values.
Based on the results, we set $\beta=0.8$ on ImageNet-R and $\beta=0.6$ on ImageNet-A.

\subsection{Experimental Results Using the CLIP Model as the Backbone}
\label{clip}
 
\begin{table*}[h]
\centering
\caption{Experimental results on ImageNet-R, CIFAR100 and CUB datasets using CLIP as the backbone. All methods are initialized with the CLIP pre-trained on LAION-400M \emph{\textbf{without experience replay}} for fair comparisons.
PROOF$^\dag$ represents our reproduced results of PROOF~\cite{PROOF-TPAMI25} without experience replay. The best results are in \textbf{bold} and the second highest results are \underline{underlined}.
}\label{tab: imagenetr_cifar_ucf}
\resizebox{\textwidth}{!}{
\begin{tabular}{@{}lccccccccccccccc@{}}
\toprule
\multicolumn{1}{c}{\multirow{3}{*}{Method}} &
\multicolumn{4}{c}{ImageNet-R} & \multicolumn{4}{c}{CIFAR100} & \multicolumn{4}{c}{CUB} \\
\cmidrule(l){2-13} 
& \multicolumn{2}{c}{B0 Inc10} & \multicolumn{2}{c}{B50 Inc10} & \multicolumn{2}{c}{B0 Inc10} & \multicolumn{2}{c}{B50 Inc10} & \multicolumn{2}{c}{B0 Inc20} & \multicolumn{2}{c}{B100 Inc20} \\
\cmidrule(l){2-13} 
& $\mathcal{A}_{Last}$ & $\mathcal{A}_{Avg}$ 
& $\mathcal{A}_{Last}$ & $\mathcal{A}_{Avg}$ 
& $\mathcal{A}_{Last}$ & $\mathcal{A}_{Avg}$ 
& $\mathcal{A}_{Last}$ & $\mathcal{A}_{Avg}$ 
& $\mathcal{A}_{Last}$ & $\mathcal{A}_{Avg}$ 
& $\mathcal{A}_{Last}$ & $\mathcal{A}_{Avg}$ \\
\midrule
CoOp~\cite{CoOp} & 37.52 & 60.73 & 39.77 & 54.20 & 24.24 & 47.00 & 24.12 & 41.23 & 8.57 & 27.61 & 10.14 & 24.03 \\
ZS-CLIP~\cite{zsclip} & 77.17 & 83.37 & 77.17 & 79.57 & 71.38 & 81.81 & 71.38 & 76.49 & 63.06 & 74.38 & 63.06 & 67.96\\
L2P~\cite{l2p} & 66.52 & 75.97 & 66.77 & 72.82 & 73.03 & 82.74 & 73.61 & 81.14 & 57.93 & 70.87 & 66.12 & 75.64\\
DualPrompt~\cite{dualprompt} & 66.65 & 76.21 & 67.58 & 73.22 & 72.44 & 81.63 & 72.57 & 80.12 & 57.46 & 69.89 & 64.84 & 71.40\\
CODA-Prompt~\cite{codaprompt} & 68.95 & 77.69 & 68.05 & 73.71 & 73.43 & 82.43 & 71.58 & 78.69 & 62.98 & 73.12 & 62.21 & 73.95\\
SimpleCIL~\cite{simplecl} & 74.48 & 81.06 & 74.48 & 76.84 & 76.63 & 84.15 & 76.63 & 80.20 & \underline{77.52} & \underline{83.81} & \underline{77.52} & \underline{79.75}\\
RAPF~\cite{rapf} & \underline{79.62} & \underline{86.28} & \underline{79.21} & \underline{84.03} & \underline{79.04} & \underline{86.19} & \underline{78.25} & \underline{82.35} & 62.77 & 79.09 & 62.93 & 72.82\\
PROOF$^\dag$~\cite{PROOF-TPAMI25} & 77.25 & 82.69 & 77.03 & 80.56 & 76.29 & 84.88 & 75.93 & 81.85 & 72.60 & 79.94 & 71.80 & 76.43\\
\midrule
\textbf{$\textbf{M}^2$LE (Ours)} & \textbf{82.05} & \textbf{87.16} & \textbf{82.65} & \textbf{85.24} & \textbf{80.52} & \textbf{87.88} & \textbf{83.12} & \textbf{87.80} & \textbf{80.03} & \textbf{86.92} & \textbf{80.41} & \textbf{84.25} \\
\bottomrule
\end{tabular}
}
\end{table*}

\begin{table*}[h]
\centering
\caption{Experimental results on Aircraft, Cars and UCF datasets using CLIP as the backbone. All methods are initialized with the CLIP pre-trained on LAION-400M \emph{\textbf{without experience replay}} for fair comparisons.
PROOF$^\dag$ represents our reproduced results of PROOF~\cite{PROOF-TPAMI25} without experience replay. The best results are in \textbf{bold} and the second highest results are \underline{underlined}.
}\label{tab: aircraft_cars_cub}
\resizebox{\textwidth}{!}{
\begin{tabular}{@{}lccccccccccccccc@{}}
\toprule
\multicolumn{1}{c}{\multirow{3}{*}{Method}} &
\multicolumn{4}{c}{Aircraft} & \multicolumn{4}{c}{Cars} & \multicolumn{4}{c}{UCF} \\
\cmidrule(l){2-13} 
& \multicolumn{2}{c}{B0 Inc10} & \multicolumn{2}{c}{B50 Inc10} & \multicolumn{2}{c}{B0 Inc10} & \multicolumn{2}{c}{B50 Inc10} & \multicolumn{2}{c}{B0 Inc20} & \multicolumn{2}{c}{B100 Inc20} \\
\cmidrule(l){2-13} 
& $\mathcal{A}_{Last}$ & $\mathcal{A}_{Avg}$ 
& $\mathcal{A}_{Last}$ & $\mathcal{A}_{Avg}$ 
& $\mathcal{A}_{Last}$ & $\mathcal{A}_{Avg}$ 
& $\mathcal{A}_{Last}$ & $\mathcal{A}_{Avg}$ 
& $\mathcal{A}_{Last}$ & $\mathcal{A}_{Avg}$ 
& $\mathcal{A}_{Last}$ & $\mathcal{A}_{Avg}$ \\
\midrule
CoOp~\cite{CoOp} & 7.14 & 14.54 & 7.77 & 13.05 & 21.65 & 36.46 & 20.87 & 37.40 & 33.46 & 47.85 & 24.74 & 42.02 \\
ZS-CLIP~\cite{zsclip} & 17.22 & 26.66 & 17.22 & 21.70 & 76.37 & 82.60 & 76.37 & 78.32 & 67.64 & 75.50 & 67.64 & 71.44 \\
L2P~\cite{l2p} & 28.29 & 47.19 & 32.13 & 44.07 & 61.82 & 76.63 & 65.64 & 76.37 & 76.43 & 86.34 & 76.62 & 83.95 \\
DualPrompt~\cite{dualprompt} & 25.83 & 44.30 & 33.57 & 46.07 & 62.94 & 76.26 & 67.55 & 76.88 & 75.82 & 85.21 & 76.35 & 84.31 \\
CODA-Prompt~\cite{codaprompt} & 27.69 & 45.98 & 32.28 & 45.14 & 66.47 & 80.21 & 64.19 & 75.06 & 80.14 & 87.76 & 75.03 & 83.04 \\
SimpleCIL~\cite{simplecl} & \underline{48.09} & \underline{59.24} & \underline{48.09} & 53.05 & \underline{86.85} & \underline{92.04} & \underline{86.85} & \underline{88.96} & \underline{85.68} & 90.44 & \underline{85.68} & 88.12 \\
RAPF~\cite{rapf} & 23.61 & 50.38 & 25.44 & 40.47 & 62.85 & 82.89 & 63.19 & 75.87 & 80.33 & \underline{92.28} & 81.55 & \underline{90.31} \\
PROOF$^\dag$~\cite{PROOF-TPAMI25} & 39.03 & 47.33 & 45.36 & \underline{54.35} & 84.06 & 86.87 & 82.66 & 88.68 & 81.47 & 87.04 & 81.85 & 86.15\\
\midrule
\textbf{$\textbf{M}^2$LE (Ours)} & \textbf{60.43} & \textbf{70.63} & \textbf{63.85} & \textbf{69.50} & \textbf{90.57} & \textbf{94.39} & \textbf{90.91} & \textbf{92.40} & \textbf{90.53} & \textbf{94.38} & \textbf{91.28} & \textbf{94.81} \\
\bottomrule
\end{tabular}
}
\end{table*}

To examine whether $\text{M}^2$LE can be transferred beyond a single visual backbone, we further conduct experiments with CLIP as the pre-trained model. 
The results on six benchmarks are summarized in Tab.~\ref{tab: imagenetr_cifar_ucf} and Tab.~\ref{tab: aircraft_cars_cub}.

\paragraph{Datasets and Evaluation Metrics.}
Following existing CLIP-based continual learning studies~\cite{PROOF-TPAMI25, rapf}, we evaluate $\text{M}^2$LE on ImageNet-R~\cite{-r}, CIFAR100~\cite{cifar100}, CUB~\cite{cub}, Aircraft~\cite{aircraft_dataset}, Cars~\cite{cars}, and UCF~\cite{ucf101_dataset}. 
ImageNet-R contains 200 ImageNet categories with different artistic renditions, such as sketches, cartoons, and paintings. 
CIFAR100 includes 100 object categories with 60,000 natural images. 
CUB, Aircraft, and Cars are fine-grained recognition datasets, covering bird species, aircraft model variants, and car categories, respectively. 
UCF101 is an action recognition benchmark with 101 human action classes. 
For Aircraft, Cars, and UCF, we follow prior protocols and sample 100 classes for continual learning evaluation. 
The split notation ``B-x Inc-y'' denotes that the first stage contains $x$ classes and each later stage introduces $y$ new classes. 
When $x=0$, all tasks contain the same number of classes.

\paragraph{Implementation Details of $\text{M}^2$LE based on CLIP.}
We build $\text{M}^2$LE upon SimpleCIL~\cite{simplecl}, a strong training-free baseline that freezes the pre-trained model and constructs a prototypical classifier from downstream features. 
Specifically, we keep one frozen CLIP branch $f_{\mathrm{pt}}(\cdot)$ together with its prototype classifier $\mathbf{W}_{\mathrm{pt}}$, and sequentially adapt another CLIP branch $f_{\mathrm{ft}}(\cdot)$ using $\text{M}^2$LE. 
The adapted branch is trained with the CLIP-style image--text matching objective~\cite{PROOF-TPAMI25}, where the text classifier is denoted as $\mathbf{W}_{\mathrm{text}}$. 
At inference time, we combine the predictions from the frozen branch and the adapted branch:
\begin{equation}
\mathbf{p} = \alpha \mathbf{W}_{\mathrm{pt}}^\top f_{\mathrm{pt}}(\mathbf{x}) + (1-\alpha) \mathbf{W}_{\mathrm{text}}^\top f_{\mathrm{ft}}(\mathbf{x}).
\end{equation}

\paragraph{Comparison Methods.}
We compare $\text{M}^2$LE with a wide range of CLIP-based continual learning baselines, including prompt-based methods CoOp~\cite{CoOp}, L2P~\cite{l2p}, DualPrompt~\cite{dualprompt}, and CODA-Prompt~\cite{codaprompt}; CLIP-oriented methods RAPF~\cite{rapf} and PROOF~\cite{PROOF-TPAMI25}; the training-free SimpleCIL~\cite{simplecl}; and the zero-shot CLIP baseline~\cite{zsclip}. 
For fair comparison, all methods use the CLIP backbone pre-trained on LAION-400M, and no experience replay is used.

\paragraph{Results and Analysis.}
As shown in Tab.~\ref{tab: imagenetr_cifar_ucf} and Tab.~\ref{tab: aircraft_cars_cub}, $\text{M}^2$LE achieves the best performance across all CLIP-based settings, showing that it can generalize well to vision-language pre-trained models. Some adaptation baselines even perform worse than ZS-CLIP or SimpleCIL, indicating that directly updating CLIP may disturb its pre-trained knowledge. In contrast, $\text{M}^2$LE enables effective layer-wise continual adaptation through adaptive LoRA merging: HIM assigns element-wise merge strengths to better balance old and new knowledge, while TSP improves the stability of the learned update for later merging. Moreover, $\text{M}^2$LE is replay-free and does not require class-wise feature statistics, making it efficient and scalable as the label space grows.

\section{Limitations}
\label{sec: limitations}

Although $\text{M}^2$LE shows strong performance across diverse continual learning benchmarks, its effectiveness still depends on the quality and adaptability of the underlying pre-trained backbone. 
When the backbone provides less transferable representations or is not well suited to low-rank updates, the benefit of LoRA-based continual adaptation may be reduced.

\section{Potential Impact}
\label{sec: society_impact}

$\text{M}^2$LE achieves efficient continual adaptation by updating only low-rank trainable parameters, which makes it suitable for long task sequences. 
However, this efficiency-oriented design may not always be optimal for accuracy-critical applications, such as medical imaging or other safety-sensitive scenarios. 
In such cases, practitioners should carefully evaluate whether the resulting performance meets the precision requirements of the target domain.

\newpage
\section*{NeurIPS Paper Checklist}

\begin{enumerate}

\item {\bf Claims}
    \item[] Question: Do the main claims made in the abstract and introduction accurately reflect the paper's contributions and scope?
    \item[] Answer: \answerYes{} 
    \item[] Justification: 
    The abstract and introduction clearly state the research objectives, methods, and contributions, all of which are consistently elaborated and supported throughout the paper and supplementary materials.
    \item[] Guidelines:
    \begin{itemize}
        \item The answer \answerNA{} means that the abstract and introduction do not include the claims made in the paper.
        \item The abstract and/or introduction should clearly state the claims made, including the contributions made in the paper and important assumptions and limitations. A \answerNo{} or \answerNA{} answer to this question will not be perceived well by the reviewers. 
        \item The claims made should match theoretical and experimental results, and reflect how much the results can be expected to generalize to other settings. 
        \item It is fine to include aspirational goals as motivation as long as it is clear that these goals are not attained by the paper. 
    \end{itemize}

\item {\bf Limitations}
    \item[] Question: Does the paper discuss the limitations of the work performed by the authors?
    \item[] Answer: \answerYes{} 
    \item[] Justification: Please see Appendix \ref{sec: limitations}.
    \item[] Guidelines:
    \begin{itemize}
        \item The answer \answerNA{} means that the paper has no limitation while the answer \answerNo{} means that the paper has limitations, but those are not discussed in the paper. 
        \item The authors are encouraged to create a separate ``Limitations'' section in their paper.
        \item The paper should point out any strong assumptions and how robust the results are to violations of these assumptions (e.g., independence assumptions, noiseless settings, model well-specification, asymptotic approximations only holding locally). The authors should reflect on how these assumptions might be violated in practice and what the implications would be.
        \item The authors should reflect on the scope of the claims made, e.g., if the approach was only tested on a few datasets or with a few runs. In general, empirical results often depend on implicit assumptions, which should be articulated.
        \item The authors should reflect on the factors that influence the performance of the approach. For example, a facial recognition algorithm may perform poorly when image resolution is low or images are taken in low lighting. Or a speech-to-text system might not be used reliably to provide closed captions for online lectures because it fails to handle technical jargon.
        \item The authors should discuss the computational efficiency of the proposed algorithms and how they scale with dataset size.
        \item If applicable, the authors should discuss possible limitations of their approach to address problems of privacy and fairness.
        \item While the authors might fear that complete honesty about limitations might be used by reviewers as grounds for rejection, a worse outcome might be that reviewers discover limitations that aren't acknowledged in the paper. The authors should use their best judgment and recognize that individual actions in favor of transparency play an important role in developing norms that preserve the integrity of the community. Reviewers will be specifically instructed to not penalize honesty concerning limitations.
    \end{itemize}

\item {\bf Theory assumptions and proofs}
    \item[] Question: For each theoretical result, does the paper provide the full set of assumptions and a complete (and correct) proof?
    \item[] Answer: \answerYes{} 
    \item[] Justification: The paper provides clearly stated assumptions and complete derivations for the theoretical results. The corresponding formulas and proofs are presented with proper references, and additional details are included in Appendix when necessary.
    \item[] Guidelines:
    \begin{itemize}
        \item The answer \answerNA{} means that the paper does not include theoretical results. 
        \item All the theorems, formulas, and proofs in the paper should be numbered and cross-referenced.
        \item All assumptions should be clearly stated or referenced in the statement of any theorems.
        \item The proofs can either appear in the main paper or the supplemental material, but if they appear in the supplemental material, the authors are encouraged to provide a short proof sketch to provide intuition. 
        \item Inversely, any informal proof provided in the core of the paper should be complemented by formal proofs provided in appendix or supplemental material.
        \item Theorems and Lemmas that the proof relies upon should be properly referenced. 
    \end{itemize}

    \item {\bf Experimental result reproducibility}
    \item[] Question: Does the paper fully disclose all the information needed to reproduce the main experimental results of the paper to the extent that it affects the main claims and/or conclusions of the paper (regardless of whether the code and data are provided or not)?
    \item[] Answer: \answerYes{} 
    \item[] Justification: Our method and experimental setup are thoroughly described in Sec.\ref{experment} to ensure reproducibility. Code for reproducing the main results is provided in the supplementary material.
    \item[] Guidelines:
    \begin{itemize}
        \item The answer \answerNA{} means that the paper does not include experiments.
        \item If the paper includes experiments, a \answerNo{} answer to this question will not be perceived well by the reviewers: Making the paper reproducible is important, regardless of whether the code and data are provided or not.
        \item If the contribution is a dataset and\slash or model, the authors should describe the steps taken to make their results reproducible or verifiable. 
        \item Depending on the contribution, reproducibility can be accomplished in various ways. For example, if the contribution is a novel architecture, describing the architecture fully might suffice, or if the contribution is a specific model and empirical evaluation, it may be necessary to either make it possible for others to replicate the model with the same dataset, or provide access to the model. In general. releasing code and data is often one good way to accomplish this, but reproducibility can also be provided via detailed instructions for how to replicate the results, access to a hosted model (e.g., in the case of a large language model), releasing of a model checkpoint, or other means that are appropriate to the research performed.
        \item While NeurIPS does not require releasing code, the conference does require all submissions to provide some reasonable avenue for reproducibility, which may depend on the nature of the contribution. For example
        \begin{enumerate}
            \item If the contribution is primarily a new algorithm, the paper should make it clear how to reproduce that algorithm.
            \item If the contribution is primarily a new model architecture, the paper should describe the architecture clearly and fully.
            \item If the contribution is a new model (e.g., a large language model), then there should either be a way to access this model for reproducing the results or a way to reproduce the model (e.g., with an open-source dataset or instructions for how to construct the dataset).
            \item We recognize that reproducibility may be tricky in some cases, in which case authors are welcome to describe the particular way they provide for reproducibility. In the case of closed-source models, it may be that access to the model is limited in some way (e.g., to registered users), but it should be possible for other researchers to have some path to reproducing or verifying the results.
        \end{enumerate}
    \end{itemize}

\item {\bf Open access to data and code}
    \item[] Question: Does the paper provide open access to the data and code, with sufficient instructions to faithfully reproduce the main experimental results, as described in supplemental material?
    \item[] Answer: \answerYes{} 
    \item[] Justification: We provide the code in the supplementary material.
    \item[] Guidelines:
    \begin{itemize}
        \item The answer \answerNA{} means that paper does not include experiments requiring code.
        \item Please see the NeurIPS code and data submission guidelines (\url{https://neurips.cc/public/guides/CodeSubmissionPolicy}) for more details.
        \item While we encourage the release of code and data, we understand that this might not be possible, so \answerNo{} is an acceptable answer. Papers cannot be rejected simply for not including code, unless this is central to the contribution (e.g., for a new open-source benchmark).
        \item The instructions should contain the exact command and environment needed to run to reproduce the results. See the NeurIPS code and data submission guidelines (\url{https://neurips.cc/public/guides/CodeSubmissionPolicy}) for more details.
        \item The authors should provide instructions on data access and preparation, including how to access the raw data, preprocessed data, intermediate data, and generated data, etc.
        \item The authors should provide scripts to reproduce all experimental results for the new proposed method and baselines. If only a subset of experiments are reproducible, they should state which ones are omitted from the script and why.
        \item At submission time, to preserve anonymity, the authors should release anonymized versions (if applicable).
        \item Providing as much information as possible in supplemental material (appended to the paper) is recommended, but including URLs to data and code is permitted.
    \end{itemize}

\item {\bf Experimental setting/details}
    \item[] Question: Does the paper specify all the training and test details (e.g., data splits, hyperparameters, how they were chosen, type of optimizer) necessary to understand the results?
    \item[] Answer: \answerYes{} 
    \item[] Justification: Please see \textbf{Datasets}, \textbf{Evaluation Metrics} and \textbf{Implementation Details} in Sec.\ref{experment}.
    \item[] Guidelines:
    \begin{itemize}
        \item The answer \answerNA{} means that the paper does not include experiments.
        \item The experimental setting should be presented in the core of the paper to a level of detail that is necessary to appreciate the results and make sense of them.
        \item The full details can be provided either with the code, in appendix, or as supplemental material.
    \end{itemize}

\item {\bf Experiment statistical significance}
    \item[] Question: Does the paper report error bars suitably and correctly defined or other appropriate information about the statistical significance of the experiments?
    \item[] Answer: \answerYes{} 
    \item[] Justification: Our experiments use three random seeds, except for the
    analysis in Table~\ref{tab:historical_fisher}, which uses a single seed. We report mean performance for the three seed experiments, and we report the average performance with standard deviation.
    \item[] Guidelines:
    \begin{itemize}
        \item The answer \answerNA{} means that the paper does not include experiments.
        \item The authors should answer \answerYes{} if the results are accompanied by error bars, confidence intervals, or statistical significance tests, at least for the experiments that support the main claims of the paper.
        \item The factors of variability that the error bars are capturing should be clearly stated (for example, train/test split, initialization, random drawing of some parameter, or overall run with given experimental conditions).
        \item The method for calculating the error bars should be explained (closed form formula, call to a library function, bootstrap, etc.)
        \item The assumptions made should be given (e.g., Normally distributed errors).
        \item It should be clear whether the error bar is the standard deviation or the standard error of the mean.
        \item It is OK to report 1-sigma error bars, but one should state it. The authors should preferably report a 2-sigma error bar than state that they have a 96\% CI, if the hypothesis of Normality of errors is not verified.
        \item For asymmetric distributions, the authors should be careful not to show in tables or figures symmetric error bars that would yield results that are out of range (e.g., negative error rates).
        \item If error bars are reported in tables or plots, the authors should explain in the text how they were calculated and reference the corresponding figures or tables in the text.
    \end{itemize}

\item {\bf Experiments compute resources}
    \item[] Question: For each experiment, does the paper provide sufficient information on the computer resources (type of compute workers, memory, time of execution) needed to reproduce the experiments?
    \item[] Answer: \answerYes{} 
    \item[] Justification: We provide the information of our compute resources in Appendix.\ref{rank}.
    \item[] Guidelines:
    \begin{itemize}
        \item The answer \answerNA{} means that the paper does not include experiments.
        \item The paper should indicate the type of compute workers CPU or GPU, internal cluster, or cloud provider, including relevant memory and storage.
        \item The paper should provide the amount of compute required for each of the individual experimental runs as well as estimate the total compute. 
        \item The paper should disclose whether the full research project required more compute than the experiments reported in the paper (e.g., preliminary or failed experiments that didn't make it into the paper). 
    \end{itemize}
    
\item {\bf Code of ethics}
    \item[] Question: Does the research conducted in the paper conform, in every respect, with the NeurIPS Code of Ethics \url{https://neurips.cc/public/EthicsGuidelines}?
    \item[] Answer: \answerYes{} 
    \item[] Justification: Our research complies the NeurIPS Code of Ethics in all aspects.
    \item[] Guidelines:
    \begin{itemize}
        \item The answer \answerNA{} means that the authors have not reviewed the NeurIPS Code of Ethics.
        \item If the authors answer \answerNo, they should explain the special circumstances that require a deviation from the Code of Ethics.
        \item The authors should make sure to preserve anonymity (e.g., if there is a special consideration due to laws or regulations in their jurisdiction).
    \end{itemize}

\item {\bf Broader impacts}
    \item[] Question: Does the paper discuss both potential positive societal impacts and negative societal impacts of the work performed?
    \item[] Answer: \answerYes{} 
    \item[] Justification: Please see Appendix \ref{sec: society_impact}.
    \item[] Guidelines:
    \begin{itemize}
        \item The answer \answerNA{} means that there is no societal impact of the work performed.
        \item If the authors answer \answerNA{} or \answerNo, they should explain why their work has no societal impact or why the paper does not address societal impact.
        \item Examples of negative societal impacts include potential malicious or unintended uses (e.g., disinformation, generating fake profiles, surveillance), fairness considerations (e.g., deployment of technologies that could make decisions that unfairly impact specific groups), privacy considerations, and security considerations.
        \item The conference expects that many papers will be foundational research and not tied to particular applications, let alone deployments. However, if there is a direct path to any negative applications, the authors should point it out. For example, it is legitimate to point out that an improvement in the quality of generative models could be used to generate Deepfakes for disinformation. On the other hand, it is not needed to point out that a generic algorithm for optimizing neural networks could enable people to train models that generate Deepfakes faster.
        \item The authors should consider possible harms that could arise when the technology is being used as intended and functioning correctly, harms that could arise when the technology is being used as intended but gives incorrect results, and harms following from (intentional or unintentional) misuse of the technology.
        \item If there are negative societal impacts, the authors could also discuss possible mitigation strategies (e.g., gated release of models, providing defenses in addition to attacks, mechanisms for monitoring misuse, mechanisms to monitor how a system learns from feedback over time, improving the efficiency and accessibility of ML).
    \end{itemize}
    
\item {\bf Safeguards}
    \item[] Question: Does the paper describe safeguards that have been put in place for responsible release of data or models that have a high risk for misuse (e.g., pre-trained language models, image generators, or scraped datasets)?
    \item[] Answer: \answerNA{} 
    \item[] Justification: Our paper poses no such risks.
    \item[] Guidelines:
    \begin{itemize}
        \item The answer \answerNA{} means that the paper poses no such risks.
        \item Released models that have a high risk for misuse or dual-use should be released with necessary safeguards to allow for controlled use of the model, for example by requiring that users adhere to usage guidelines or restrictions to access the model or implementing safety filters. 
        \item Datasets that have been scraped from the Internet could pose safety risks. The authors should describe how they avoided releasing unsafe images.
        \item We recognize that providing effective safeguards is challenging, and many papers do not require this, but we encourage authors to take this into account and make a best faith effort.
    \end{itemize}

\item {\bf Licenses for existing assets}
    \item[] Question: Are the creators or original owners of assets (e.g., code, data, models), used in the paper, properly credited and are the license and terms of use explicitly mentioned and properly respected?
    \item[] Answer: \answerYes{} 
    \item[] Justification: 
    The existing assets covered in the paper are subject to their license and terms of use, and are explicitly cited and described.
    \item[] Guidelines:
    \begin{itemize}
        \item The answer \answerNA{} means that the paper does not use existing assets.
        \item The authors should cite the original paper that produced the code package or dataset.
        \item The authors should state which version of the asset is used and, if possible, include a URL.
        \item The name of the license (e.g., CC-BY 4.0) should be included for each asset.
        \item For scraped data from a particular source (e.g., website), the copyright and terms of service of that source should be provided.
        \item If assets are released, the license, copyright information, and terms of use in the package should be provided. For popular datasets, \url{paperswithcode.com/datasets} has curated licenses for some datasets. Their licensing guide can help determine the license of a dataset.
        \item For existing datasets that are re-packaged, both the original license and the license of the derived asset (if it has changed) should be provided.
        \item If this information is not available online, the authors are encouraged to reach out to the asset's creators.
    \end{itemize}

\item {\bf New assets}
    \item[] Question: Are new assets introduced in the paper well documented and is the documentation provided alongside the assets?
    \item[] Answer: \answerYes{} 
    \item[] Justification: 
    We provide the code in the supplementary material.
    \item[] Guidelines:
    \begin{itemize}
        \item The answer \answerNA{} means that the paper does not release new assets.
        \item Researchers should communicate the details of the dataset\slash code\slash model as part of their submissions via structured templates. This includes details about training, license, limitations, etc. 
        \item The paper should discuss whether and how consent was obtained from people whose asset is used.
        \item At submission time, remember to anonymize your assets (if applicable). You can either create an anonymized URL or include an anonymized zip file.
    \end{itemize}

\item {\bf Crowdsourcing and research with human subjects}
    \item[] Question: For crowdsourcing experiments and research with human subjects, does the paper include the full text of instructions given to participants and screenshots, if applicable, as well as details about compensation (if any)? 
    \item[] Answer: \answerNA{} 
    \item[] Justification: 
    This paper does not involve crowdsourcing nor research with human subjects.
    \item[] Guidelines:
    \begin{itemize}
        \item The answer \answerNA{} means that the paper does not involve crowdsourcing nor research with human subjects.
        \item Including this information in the supplemental material is fine, but if the main contribution of the paper involves human subjects, then as much detail as possible should be included in the main paper. 
        \item According to the NeurIPS Code of Ethics, workers involved in data collection, curation, or other labor should be paid at least the minimum wage in the country of the data collector. 
    \end{itemize}

\item {\bf Institutional review board (IRB) approvals or equivalent for research with human subjects}
    \item[] Question: Does the paper describe potential risks incurred by study participants, whether such risks were disclosed to the subjects, and whether Institutional Review Board (IRB) approvals (or an equivalent approval/review based on the requirements of your country or institution) were obtained?
    \item[] Answer: \answerNA{} 
    \item[] Justification: 
    This paper does not involve crowdsourcing nor research with human subjects.
    \item[] Guidelines:
    \begin{itemize}
        \item The answer \answerNA{} means that the paper does not involve crowdsourcing nor research with human subjects.
        \item Depending on the country in which research is conducted, IRB approval (or equivalent) may be required for any human subjects research. If you obtained IRB approval, you should clearly state this in the paper. 
        \item We recognize that the procedures for this may vary significantly between institutions and locations, and we expect authors to adhere to the NeurIPS Code of Ethics and the guidelines for their institution. 
        \item For initial submissions, do not include any information that would break anonymity (if applicable), such as the institution conducting the review.
    \end{itemize}

\item {\bf Declaration of LLM usage}
    \item[] Question: Does the paper describe the usage of LLMs if it is an important, original, or non-standard component of the core methods in this research? Note that if the LLM is used only for writing, editing, or formatting purposes and does \emph{not} impact the core methodology, scientific rigor, or originality of the research, declaration is not required.
    \item[] Answer: \answerNA{} 
    \item[] Justification: 
    No, our paper does not involve the use of large language models (LLMs) as a component of the core methodology. 
    \item[] Guidelines:
    \begin{itemize}
        \item The answer \answerNA{} means that the core method development in this research does not involve LLMs as any important, original, or non-standard components.
        \item Please refer to our LLM policy in the NeurIPS handbook for what should or should not be described.
    \end{itemize}

\end{enumerate}

\end{document}